\documentclass[twoside,11pt]{article}
\usepackage[preprint]{jmlr2e}
\usepackage{url}            
\usepackage{hyperref}
\usepackage{booktabs}       
\usepackage{amsfonts}       
\usepackage{amsmath}
\usepackage{algorithm}
\usepackage{algpseudocode}
\usepackage{mathrsfs}
\usepackage{utfsym}
\usepackage{nicefrac}       
\usepackage{microtype}      
\usepackage{xcolor}         
\usepackage{color}
\usepackage{graphicx}
\usepackage{enumitem}
\usepackage{caption}
\usepackage{subcaption}
\usepackage{multirow}
\usepackage{mathtools}

\newcommand{\zl}[1]{\textcolor{teal}{Zongyi: #1}}

\usepackage{lastpage}
\jmlrheading{26}{2025}{1-\pageref{LastPage}}{5/21}{ / }{21-0000}{Xinyi Li and Zongyi Li and Nikola Kovachki and Anima Anandkumar}

\ShortHeadings{Geometric Operator Learning with Optimal Transport}{Xinyi Li and Zongyi Li and Nikola Kovachki and Anima Anandkumar}
\firstpageno{1}

\begin{document}

\title{Geometric Operator Learning with Optimal Transport}

\author{\name Xinyi Li 
\email xinyili@caltech.edu \\
       \addr 
       California Institute of Technology
       \AND
       \name Zongyi Li \email zongyili@caltech.edu \\
       \addr
       California Institute of Technology
       \AND
       \name Nikola Kovachki \email nkovachki@nvidia.com \\
       \addr
       Nvidia
       \AND
       \name Anima Anandkumar  \email anima@caltech.edu \\
       \addr
       California Institute of Technology
}
\editor{My editor}

\maketitle

\begin{abstract}
We propose integrating optimal transport (OT) into operator learning for partial differential equations (PDEs) on complex geometries. Classical geometric learning methods typically represent domains as meshes, graphs, or point clouds. Our approach generalizes discretized meshes to mesh density functions, formulating geometry embedding as an OT problem that maps these functions to a uniform density in a reference space. Compared to previous methods relying on interpolation or shared deformation, our OT-based method employs instance-dependent deformation, offering enhanced flexibility and effectiveness. For 3D simulations focused on surfaces, our OT-based neural operator embeds the surface geometry into a 2D parameterized latent space. By performing computations directly on this 2D representation of the surface manifold, it achieves significant computational efficiency gains compared to volumetric simulation. Experiments with Reynolds-averaged Navier-Stokes equations (RANS) on the ShapeNet-Car and DrivAerNet-Car datasets show that our method achieves better accuracy and also reduces computational expenses in terms of both time and memory usage compared to existing machine learning models. Additionally, our model demonstrates significantly improved accuracy on the FlowBench dataset, underscoring the benefits of employing instance-dependent deformation for datasets with highly variable geometries.
\end{abstract}

\begin{keywords}
  Geometric Learning, Neural Operator, Optimal Transport, Computational Fluid Dynamics.
\end{keywords}
\section{Introduction}
Handling complex geometric domains in 3D space remains one of the fundamental challenges in scientific computing. While standard numerical solvers based on finite element or spectral methods have been successful on simple regular domains, they struggle with complex geometries due to computationally expensive meshing processes that often require iterative refinement.
The challenge of geometric modeling poses a obstacle across multiple domains, including fluid dynamics, solid mechanics, and earth science applications.
This challenge is particularly evident in 3D aerodynamic simulations of automobiles, where a single shape takes over three hundred hours on CPU \citep{elrefaie2024drivaernetparametriccardataset} or ten hours on GPU \citep{GINO}.


Machine learning methods have emerged as promising alternatives for solving PDEs on complex geometries, offering dramatic improvements in computational efficiency \citep{bhatnagar2019prediction, pfaff2020learning, thuerey2020deep, hennigh2021nvidia}. These approaches can operate effectively at lower resolutions compared to traditional numerical solvers, significantly reducing computational overhead. However, most existing ML-based methods are constrained to specific resolutions, limiting their flexibility and broader applicability. To address this limitation, we focus on neural operators, a recent breakthrough in scientific computing that offers a resolution-independent approach to solving PDEs.

\paragraph{Neural operators for complex geometries.}
Neural operators represent an innovative class of data-driven models designed to directly learn the mapping of solution operators for PDEs in a mesh-free manner \citep{FNO, kovachki2023neural, lu2021learning}. Unlike conventional deep learning models, neural operators are designed to be invariant to discretization, making them particularly effective for solving PDEs. 
Recent advances in neural operator research have focused on addressing PDEs with complex geometries \citep{Geo-FNO,yin2024dimon,ahmad2024diffeomorphiclatentneuraloperators}, primarily through embedding the geometries into uniform latent spaces where efficient spectral methods such as the Fast Fourier Transform (FFT) \citep{FFT} can be applied.

The Geometry-Aware Fourier Neural Operator (Geo-FNO) \citep{Geo-FNO} introduces an important technique of constructing diffeomorphic mapping from physical to computational domains structured as regular grids.  This innovation enabled the application of the FFT in latent computational spaces, dramatically improving computational efficiency. However, Geo-FNO faces two major limitations. 
It learns a shared deformation map for a class of shapes, which cannot address case-dependent geometry features well.
In addition, GeoFNO encodes and decodes the geometry using a Fourier transform based on computationally expensive matrix-vector multiplication, which restricts scaling to large 3D simulations. 

Building on these ideas, the Geometry-Informed Neural Operator (GINO) \citep{GINO} combined Graph Neural Operators (GNO) \citep{li2020neural} with Fourier Neural Operators (FNO) \citep{FNO}. By leveraging the adaptability of graphs for local interactions and the computational efficiency of FFT for global physics, it became the first neural operator capable of tackling large-scale 3D aerodynamics challenges. Despite its potential, the method struggles with the inherent limitations of graph embeddings' locality and the high computational cost of 3D latent spaces. These challenges are especially evident in large-scale scenarios.

Another line of works encode the geometry into structure-free tokens based on transformer \citep{hao2023gnot, wu2024transolver, alkin2024universal} or implicit neural representation \citep{yin2022continuous, serrano2023operator, chen2023implicit, chen2022crom}. These methods are flexible and generic, but they generally do not preserve geometric properties in their encoding. Meanwhile, their encoders are generally not invertible, which limits their applications in inverse meshing optimization and shape design.

Despite these advances, current neural operator approaches continue to grapple with the computational burden of 3D PDEs and the challenge of learning operators across diverse geometries. To address these challenges, we propose reformulating geometry embedding as an optimal transport problem for each instance. This allows solution operators to be learned directly on the surface manifold, with deformation tailored to each instance. This novel approach fundamentally transforms the handling of complex geometries in neural operators.


\paragraph{Geometry encoding with optimal transport}
Optimal transport provides a rigorous mathematical framework for determining the most efficient transformation between densitys. We leverage this framework by interpreting surface meshes as continuous density functions, where mesh density reflects the underlying geometric complexity and surface curvature. Our key insight is to formulate the geometry embedding problem as an optimal transport problem that maps these mesh density functions to uniform density functions in a latent space.

\begin{figure}[t]
    \centering
    \includegraphics[width=1\linewidth]{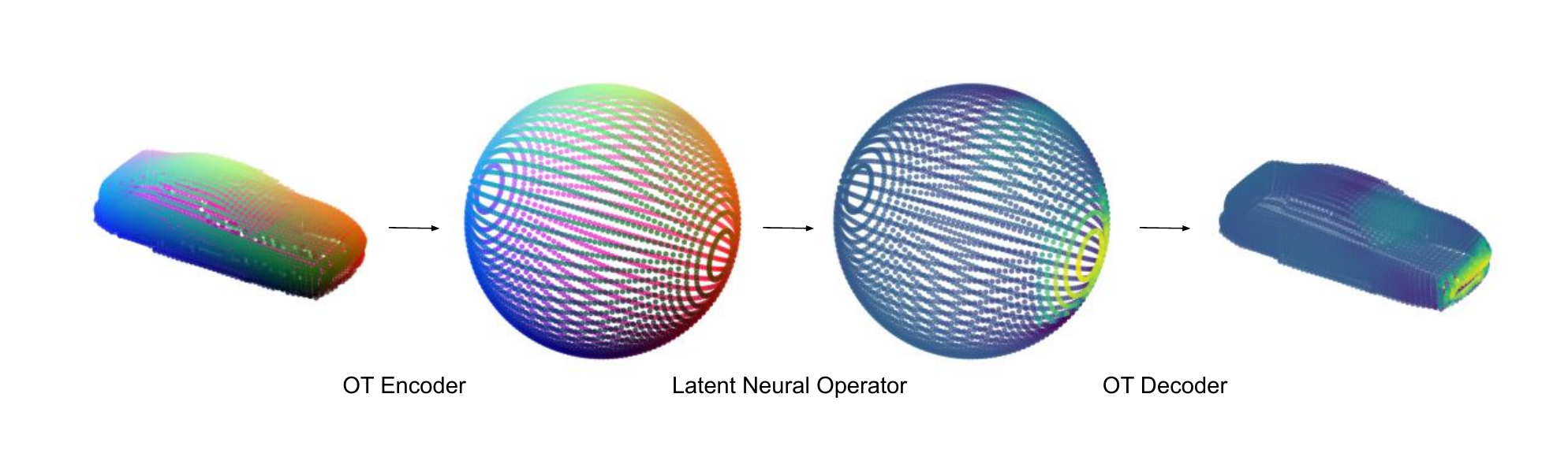}
    \caption{Illustration of the optimal transport neural operator (OTNO). (a) \textbf{OT Encoder}: The surface mesh is encoded onto a latent computational mesh, and the OT coupling is visualized by representing the coordinates of points as RGB colors. (b) \textbf{Latent Neural Operator}: Within the latent space, we apply S/FNO to calculate solutions on the latent mesh. (c) \textbf{OT Decoder}: We decode the solutions from the latent space back to the original surface mesh; here, the colors indicate solution values.
    }
    \label{fig:structure}
\end{figure}

\begin{figure}[t]
\centering
    \includegraphics[width=\textwidth]{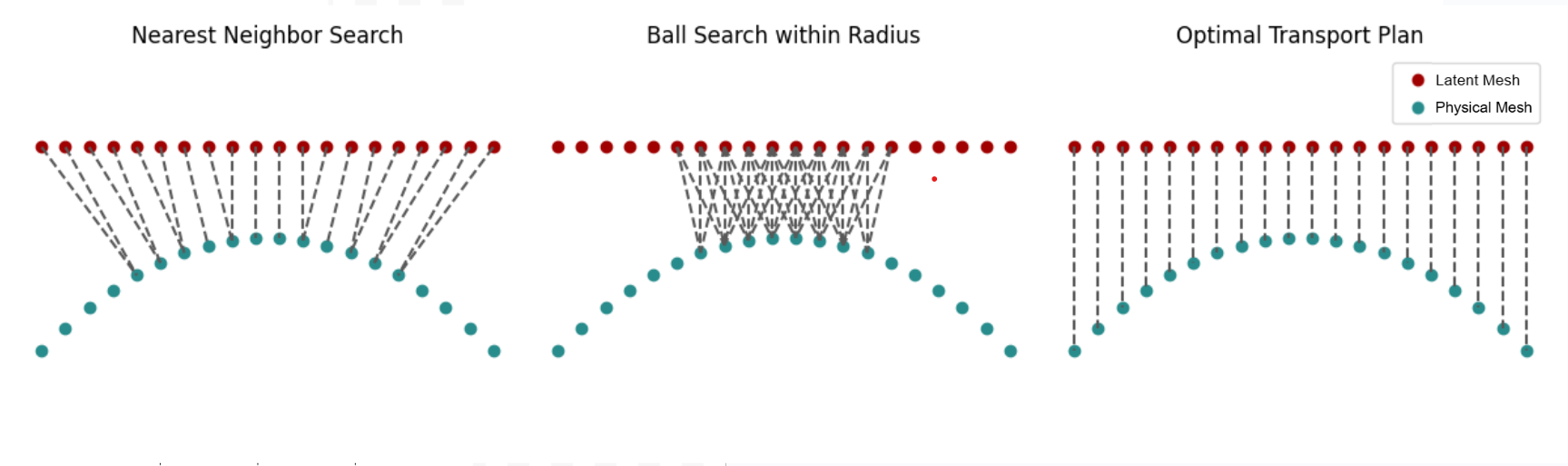} 
    \caption{OT plans can be viewed as bi-partite graphs. In the figure, the green nodes represent the input shape and the red nodes represent the latent grid. Compared to other graph mapping strategies such as ball connection and nearest neighbor connection, OT preserves the global measure, which is essential for computing integral operators.
    }
    \label{fig:graph_mapping}
\end{figure}

Unlike traditional approaches that rely on direct projection and interpolation—which often result in problematic point clustering and density distortions—optimal transport inherently preserves the structural properties of the mesh while ensuring a smooth, physically meaningful transformation, as illustrated in Figure \ref{fig:graph_mapping}. This preservation property shares conceptual similarities with adaptive moving mesh methods \citep{budd2015geometry}, but offers flexibility for different topologies. Recent computational advances, particularly the Sinkhorn algorithm \citep{cuturi2013sinkhorn}, have made optimal transport practically feasible by providing efficient approximations to the transport problem. Building on these developments, we demonstrate how optimal transport can effectively embed surface mesh sub-manifolds into latent space while maintaining their essential geometric properties, as illustrated in Figure \ref{fig:structure}.

\paragraph{Our Contribution:} 
In this work, we generalize geometry learning from discretized mesh points to mesh density function. Our key innovation lies in formulating geometry embedding as a pre-determined transform that maps the input mesh density function to the uniform density function in the canonical reference space. Such geometric transform is computed as optimal transport.  

We explore both the Monge formulation (transport maps) and Kantorovich formulation (transport plans), showing how this unified framework naturally encompasses previous approaches: transport maps generalize the deformation maps in Geo-FNO, while transport plans extend the graph representations in GINO.

Building on this theoretical foundation, we introduce the optimal transport neural operator (OTNO), which combines OT-based geometry encoding/decoding with (Spherical) Fourier Neural Operators \citep{FNO, bonev2023spherical} (Figure \ref{fig:structure}).  We use Kantorovich formulation to achieve sparse transport plan and implement it with Sinkhorn algorithm\citep{cuturi2013sinkhorn}, and use Monge formulation to obtain bijective map and implement it with Projection pursuit Monge map (PPMM)\citep{meng2019large}. 

Our optimal transport technique enables crucial dimension reduction by embedding the surface manifolds from the $d$-dimensional ambient space into the $(d-1)$-dimensional latent spaces. This capability is particularly valuable for automotive and aerospace applications, where many critical simulations, including Reynolds-averaged Navier-Stokes (RANS) and Large-Eddy Simulation (LES), fundamentally operate on boundary value problems. The input is a 2D surface design, and the desired outputs are surface quantities like pressure and shear velocity that determine total drag.

\begin{figure}[t]
    \centering
    \includegraphics[width=0.8\linewidth]{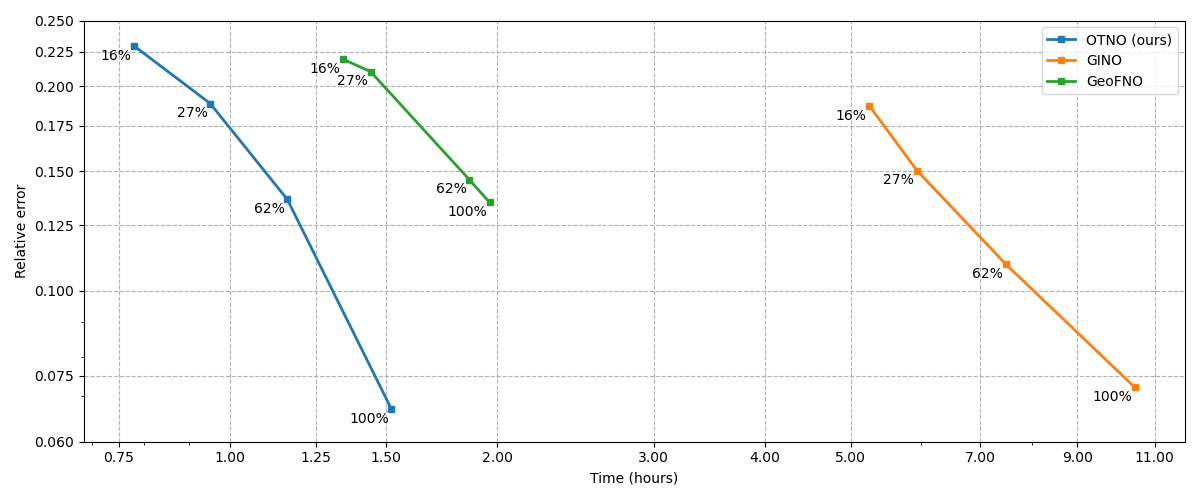}
    \caption{Convergence Plot: This figure presents a comparison of convergence rates among different models. Note that ‘Time’ denotes the total runtime, including OT computation for OTNO and SDF computation for GINO. We vary the physical mesh size by downsampling to 10\%, 27\%, and 62\% of the original points, and scale the latent mesh size proportionally (as shown in Table \ref{tab:resoluion-settings}). The data indicates that our model, OTNO, exhibits the fastest exponential convergence rate of 1.85, surpassing both GINO at 1.37 and GeoFNO at 1.32.}
    \label{fig:convergence}
\end{figure}

We validate our method through simulations based on RANS equations on the ShapeNet-Car \citep{umetani2018learning} and DrivAerNet-Car \citep{elrefaie2024drivaernet++} datasets. Results under different sampling rates (Figure \ref{fig:convergence}) show that our approach achieves the fastest convergence rate compared to baseline methods and achieves both the smallest error and shortest time cost when using full dataset. Given different sampling size, our model maintains robust performance in geometry representation and embedding.

Moreover, our OTNO conducts geometry embedding for each shape individually, distinguishing it from previous methods such as GeoFNO and GINO, which learned a shared deformation network across all geometries. As confirmed by our experiments on the FlowBench dataset, this feature enables our model to better handle datasets composed of a wider variety of shapes.
Our main contributions are summarized as follows:


\begin{enumerate} 
\item  A novel optimal transport framework for mesh embedding that unifies and generalizes previous approaches by mapping mesh density functions to latent uniform density functions, bridging the gap between Geo-FNO and GINO methodologies.
\item The Sub-Manifold Method, a dimension-reduction technique for high-dimensional PDEs that restricts solution operators to surface manifolds with with one lower dimension, implemented with optimal transport and coupled with latent spectral neural operators for efficient PDE resolution in reduced dimensional space, which achieve significant reduction in computational expense.
\item Comprehensive validation on industry-standard datasets—ShapeNet-Car (3.7k points) \citep{chang2015shapenet} and DrivAerNet-Car (400k points) \citep{elrefaie2024drivaernetparametriccardataset}—demonstrates unprecedented efficiency in RANS pressure field prediction. Our method achieves a performance improvement of 2x-8x faster processing and 2x-8x smaller memory usage compared to current machine learning methods, while also slightly enhancing accuracy. Moreover, it is approximately 7,000 times faster than traditional approaches.
\item Optimal transport provides instance-dependent geometry embeddings. Our model notably excels across diverse geometries, as evidenced on the FlowBench dataset, which contains shapes with greater variability.
\end{enumerate}
\section{Problem Setting and Preliminaries}
\label{problem setting}
\subsection{Problem settings}
In this work, we consider boundary solution problems arising from PDEs. Our aim is to learn the operator from the boundary geometries of PDEs to their boundary solutions. 
In previous works, the boundary shapes have been parameterized as discrete design parameters \citep{timmer2009overview}, occupancy functions or signed distance functions \citep{GINO}.
In this work, we model the geometries as supports of mesh density functions defined on an ambient Euclidean space. To make this concrete, we consider first the set of all probability densities on $\mathbb{R}^d$ defined with respect to the Lebesgue measure:
\[\mathcal{F} = \{f \in L^1(\mathbb{R}^d) : f \geq 0, \|f\|_{L^1} = 1\}.\]
In this work, we will be interested in PDE problems on bounded domains. We therefore define the following restriction to the set of densities
\[\mathcal{F}_c = \{f \in \mathcal{F} : \text{supp}(f) \subset \mathbb{R}^d \text{ is a $d$-dimensional bounded manifold with boundary}\}.\]
For any density $f \in \mathcal{F}_c$, we use the notation $\Omega_f \coloneqq \text{supp}(f)$ to denote the $d$-dimensional manifold 
defined by its support. Furthermore, we use the notation $\partial \Omega_f$ to denote the boundary of $\Omega_f$ which is a $(d-1)$-dimensional sub-manifold.
Let $\mathcal{L}$ and $\mathcal{B}$ be two, possibly non-linear, partial differential operators and consider the PDE
\begin{equation}
\label{primary PDE}
\begin{aligned}
    \mathcal{L}(u) &= h, &&\text{in } \Omega_f ,\\
    \mathcal{B} (u) &= b, &&\text{in } \partial \Omega_f,
\end{aligned} 
\end{equation}
where $h$ and $b$ are some fixed functions on $\mathbb{R}^d$. For any open set $A \subseteq \mathbb{R}^d$, let $\mathcal{U}(A)$ be a Banach function space whose elements have domain $A$. We will assume that there exists a family of extension operators $E_A : \mathcal{U}(A) \to \mathcal{U}(\mathbb{R}^d)$ and fix such a family. Explicit constructions of such bounded, linear operators are available, for example, when $\mathcal{U}$ is a Lebesgue or Sobolev space and $A$ satisfies a cone condition \citep{stein1970singular}. 
We now consider the following set of densities:
\[\mathcal{F}_c^{PDE} = \mathcal{F}_c^{PDE}(\mathcal{L}, \mathcal{B}, h, b) = \{f \in \mathcal{F}_c : \exists ! u  \in \mathcal{U}(\Omega_f) \text{ s.t. } u \text{ satisfies } \eqref{primary PDE}\}.\]
We can thus define a solution operator which maps a density $f$, defining the domain of the PDE, to a uniquely extended solution of the PDE. In particular,
\begin{equation}
\begin{aligned}
    \label{eq:manifold-problem}
    \mathcal{G}^{\dagger} : \mathcal{F}_c^{PDE} \subset L^1 (\mathbb{R}^d) &\to \mathcal{U}(\mathbb{R}^d), \\
    f &\mapsto E_{\Omega_f} (u),
\end{aligned}
\end{equation}
where $u \in \mathcal{U}(\Omega_f)$ is the unique solution to \eqref{primary PDE}. Our aim is to approximate $\mathcal{G}^{\dagger}$ or the associated sub-manifold mapping defined subsequently from data. While the extension operators allow us to define $\mathcal{G}^{\dagger}$ as a mapping between two function spaces with the same domain, in practice, we will only approximate each output of $\mathcal{G}^{\dagger}$ on its domain $\Omega_f$ as this captures all relevant information about the PDE \eqref{primary PDE}. We outline this formulation, however, as we believe it is more amenable to potential theoretical analysis. 

\paragraph{Sub-Manifold Solution Operator}
For the associated boundary solution problems, the solution operator on the sub-manifold \(\partial \Omega_f\) is given by:
\begin{equation}
\label{eq:sub-manifold problem}
    \mathcal{G}^{\dagger}: f^{\text{sub}} \mapsto u^{\text{sub}} \quad \text{in} \ \partial \Omega_f,
\end{equation}
where \(f^{\text{sub}}= \frac{f|_{\partial \Omega_f}}{\int_{\partial \Omega_f} f(x)dS }\), with $dS$ a surface measure, denotes the normalized function of \(f\) constrained to \(\partial \Omega_f\) which is a density function on \(\partial \Omega_f\), and \(u^{\text{sub}}\) denotes our target solution function on \(\partial \Omega_f\). 

For linear elliptic PDEs such as the Helmholtz equations on regular domains  \citep{Helmholtz, hubbert1956darcy}, solution operators can be explicitly constructed on boundary sub-manifolds through a well-established process: defining basis functions for boundary inputs and solutions via singular value decomposition (SVD), then establishing a linear mapping between these bases. However, this approach breaks down for nonlinear problems like RANS and LES with complex geometries, where explicit linear mappings become mathematically impossible. To overcome this limitation, we introduce a novel approach that combines optimal transport for geometry embedding with neural operators learned directly on the surface sub-manifold, enabling efficient handling of nonlinear boundary problems.
\begin{figure}
    \centering
    \includegraphics[width=0.9\linewidth]{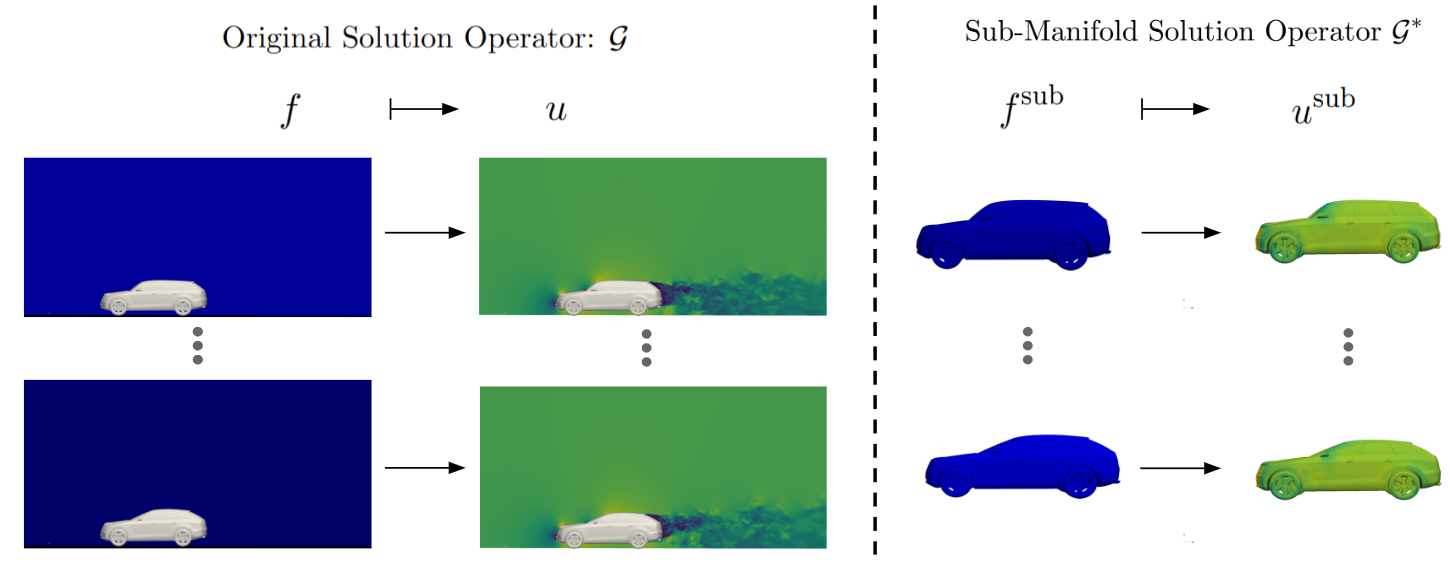}
    \caption{Illustration of Sub-Manifold Solution Operator for RANS Equation on Automotive Surfaces. The function $f$ and $f^{\text{sub}}$ are visualized using RGB colors set to $[0,0,b]$ where $b\propto \frac{1}{n}$ and $n$ is the number of mesh vertices. \textbf{Right side}: The original solution operator defined on the closed volume outside of automotive. The functions $f$ and $u$ are shown using their slices at $y=0$. And for solution function $u=(\overline{\mathbf{v}}, \overline{\mathbf{p}})$, we specifically display one component—velocity in the x-direction. \textbf{Left}: The sub-manifold solution operator is defined on the automotive surface. The solution function $u^{\text{sub}} = \overline{\mathbf{p}}$ is visualized. }
    \label{fig:sub-manifold operator}
\end{figure}
\paragraph{ Reynolds-Averaged Navier-Stokes Equations}
One common example in computational fluid dynamics is the shape design problem. Given a mesh with mesh density function $f \in V \subset \mathbb{R}^3$, we aim to solve the Reynolds-averaged Navier–Stokes Equations in $\Omega_f = \operatorname{supp}(f)$, which is the closed volume outside of automotive or airfoil. The boundary $\Omega_f$ is defined as $\partial \Omega_f = P \sqcup Q$, where $P$ denotes the far-field (or outer) boundary extending to infinity, and $Q$ represents the solid surface of automotive or airfoil (which may contain minor non-manifold structures but can be approximated as a manifold).
\begin{equation}
\begin{aligned}
  - \frac{1}{Re} \Delta \overline{\mathbf{v}} + (\overline{\mathbf{v}} \cdot \nabla) \overline{\mathbf{v}} + \nabla \overline{\mathbf{p}}  &= \mathbf{h} &&  \text{in } \Omega_f^{\circ}, \\
  \nabla \overline{\mathbf{v}} &= 0 &&  \text{in } \Omega_f^{\circ}, \\  
  \overline{\mathbf{v}} &= \mathbf{b} &&  \text{in } P,
\end{aligned}
\end{equation}

where $\overline{\mathbf{v}}$ represents the time-averaged velocity field, $\mathbf{b}$ is boundary condition of velocity, and $\overline{\mathbf{p}}$ denotes the time-averaged pressure field. The $Re$ represents the Reynolds number, which accounts for turbulence effects in the averaged flow field. The vector $\mathbf{h}$ includes any external forces acting on the fluid. Given that the boundary condition \(\mathbf{b}\) and vector \(\mathbf{h}\) are fixed, these equations on various manifolds give rise to the solution operator \( \mathcal{G}: f \mapsto u=(\overline{\mathbf{v}}, \overline{\mathbf{p}}) \).

Our objective is to solve the pressure field $\overline{\mathbf{p}}$ on the boundary manifold \(Q\) which is the surface of automotive or airfoil in practice. Correspondingly, the solution operator we target is a sub-manifold solution operator of the original 3D PDE solution operator \( \mathcal{G}\):
\begin{equation}
    \mathcal{G}^{\dagger}: f^{\text{sub}} \mapsto u^{\text{sub}}=\overline{\mathbf{p}} \quad \text{in } Q. 
\end{equation} 
where \(f^{\text{sub}}\) is the density function of the surface mesh of the automotive or airfoil, which is a mass function on it in practice. Figure~\ref{fig:sub-manifold operator} provides an illustration of this sub-manifold operator, using simulation data from the SHIFT-SUV Sample dataset \citep{shift_suv_2025}. Although this setting presents certain limitations, it remains a general framework applicable to many realistic 3D geometry design problems, where the desired solutions are typically concentrated on the surfaces of objects such as cars and airfoils. 

\subsection{Neural Operator}
Suppose there is a set of function set $\mathcal{F}$, and for each function $f\in \mathcal{F}$, it determines a specific PDE with a specific solution function $u\in \mathcal{U}$. The target is to learn the solution operator for a family of PDEs $\mathcal{G}^{\dagger}: f \mapsto u$. 

The neural operator $\mathcal{G}_{\theta}$ proposed by \cite{kovachki2023neural} composes point-wise and integral layers to approximate the target operator:

\begin{equation*}
    \mathcal{G}_{\theta}=\mathcal{Q} \circ \mathcal{K}_{L}\circ \cdots \circ \mathcal{K}_1 \circ \mathcal{P}
\end{equation*}
where $\mathcal{Q}$ and $\mathcal{P}$ are pointwise neural network that lifting the lower dimension input to higher dimension latent space $D$ and project them back to lower dimension output respectively. $\mathcal{K}_l$ is integral operator $\mathcal{K}_l:v_{l-1} \mapsto v_l$: 
\begin{equation*}
    v_l (x) = \int_D \kappa_l(x, y) v_{l-1}(y) \, \mathrm{d}y
\end{equation*}
where $\kappa_l$ is a learnable kernel function.

Usually we assume the dataset $\{f_j, u_j\}_{j=1}^N $ is available, where $\mathcal{G}(f_j)=u_j$. Then we can optimize the neural operator by minimizing the empirical data loss:

\begin{equation*}
    Loss(\mathcal{G}_\theta) = \frac{1}{N} \sum_{j=1}^n l(u_j, \mathcal{G}_\theta (f_j) ) 
\end{equation*}

where $l(\cdot,\cdot)$ denotes an appropriate error metric (e.g., MSE or relative $L^2$ error). In this work, we propose to 

\subsection{Neural Operator on Geometric Problems}
It is a standard method to embed the varying physical domain $\Omega$ into a latent space $\Omega^*$ with various types of encoders and decoders
\begin{equation}
    \mathcal{G}^{\dagger} \approx \mathcal{Q} \circ \mathcal{G}^* \circ \mathcal{P}
\end{equation}
where $\mathcal{G}^*$ is a latent operator defined on the latent space $\Omega^*$. Previous works have explored using interpolation and deformation as the encoders and decoders.

\paragraph{Geometric Informed Neural Operator}
Geometric Informed Neural Operator (GINO) \cite{GINO} assign a uniform latent grid. It defines encoder $\mathcal{Q}$ and decoder $\mathcal{P}$ as graph neural operator \cite{GNO} to learn non-linear interpolation from the physical mesh to the latent grid. 
\begin{equation*}
    v_l (x) = \sum_{y\in\mathcal{N}} \kappa_l(x, y) v_{l-1}(y) \, \mu(y)
\end{equation*}
where $\mathcal{N}$ is the neighbor of the bipartite graph between the physical mesh and the latent grid.

\paragraph{Geometric Aware Fourier Neural Operator}
Geometric Aware Fourier Neural Operator (GeoFNO) assigns a canonical manifold as the latent space, and constructs an invertible deformation map (diffeomorphism) as the encoder
\begin{equation}
    T:\Omega^* \to \Omega 
\end{equation} 
The deformation map induces a pullback of input function $a$ on the latent space, $T^\# a := a \circ T$. Then GeoFNO applies the latent operator to the pullback $T^\# a$. 

In this work, we will investigate using optimal transport as the encoder and decoder to embed the geometries into a uniform mesh.

\subsection{Optimal Transport}
Monge originally formulated the OT problem as finding the most economical map to transfer
one measure to another \citep{monge1781memoire}. Later, Kantorovich introduced a relaxation of
these strict transportation maps to more flexible transportation plans, solved using linear
programming techniques \citep{kantorovich2006problem}.

\subsubsection{Monge Problem}
\label{monge problem}
Let $\Omega$ and $\Omega^*$ be two separable metric spaces such that any probability measure on $\Omega$ (or $\Omega^*$) is a Radon measure (i.e. they are Radon spaces). Let \( c: \Omega^* \times \Omega \rightarrow \mathbb{R}^+ \) be a Borel-measurable function. Given probability measures $\mu$ on $\Omega$ and $\lambda$ on $\Omega^*$ with corresponding density functions $f$ and $g$, Monge's formulation of the optimal transportation problem is to find a transport map \(T : \Omega^* \rightarrow \Omega\) that minimizes the total transportation cost:
\begin{equation}
\label{eq: MP}
\text{(MP)} \quad \inf \{ \int_{\Omega^*} c(\xi, T(\xi)) d\lambda(\xi) \mid T_\# \lambda = \mu \} ,
\end{equation}
where  \(T_\# \lambda = \mu\) denotes that map \(T \) is measure preserving (i.e. $\int_{T^{-1}(B)} d\lambda(\xi) = \int_B d\mu(x)$, for any Borel set \(B \subseteq \Omega\) ).

Moreover, the following theorem and continuity property hold for the Monge formulation.
\begin{theorem}[Existence and uniqueness of transport map \citep{brenier1991polar}]
\label{thm:brenier}
Suppose the measures \(\mu\) and \(\lambda\) are with compact supports \(\Omega, \Omega^* \subseteq \mathbb{R}^d\) respectively, and they have equal total mass \(\mu(\Omega) = \lambda(\Omega^*)\). Assume the corresponding density functions satisfy \(f, g \in L^1(\mathbb{R}^d)\), and the cost function is \(c(\xi, x) = \frac{1}{2} |\xi - x|^2\), then the OT map from \(\lambda\) to \(\mu\) exists and is unique. It can be expressed as \( T(\xi) = \xi + \nabla \phi(\xi) \), where \(\phi : \Omega^* \rightarrow \mathbb{R}\) is a convex function, and \(\phi\) is unique up to adding a constant. 
\end{theorem}

\begin{lemma}[Continuity of transport map]
    Given that cost function is the squared Euclidean distance and $\lambda$ is a measure with uniform density function with a compact support, if $\mu$ is absolutely continuous and strictly positive also with a compact support, then the OT map $T$ is continuous almost everywhere. (This lemma can be easily derived from Theorem~\ref{thm:brenier}.)
\end{lemma}

In practical applications, especially in computational settings, the continuous problem \eqref{eq: MP} is often discretized. Suppose the measures \(\lambda\) and \(\mu\) are supported on finite point sets \(\Xi = \{ \xi_1, \dots, \xi_{n_1} \} \subset \Omega^*\) and \(\mathcal{X} = \{ x_1, \dots, x_{n_2} \} \subset \Omega\), and are represented by discrete probability vectors \(a = (\lambda_1, \dots, \lambda_{n_1})\) and \(b = (\mu_1, \dots, \mu_{n_2})\), respectively. 

Then Monge problem then seeks a transport map \(T\) that minimizes the following total cost:
\begin{equation}
    \min_{T\in \Psi} \sum_{i=1}^{n_1} \lambda_i \cdot c(\xi_i, T(\xi_i)) ,
\end{equation}
where $\Psi = \{ T \, | \, T_\# \lambda = \mu \}$ denotes the set of all feasible transport maps.

\subsubsection{Kantorovich Problem}
\label{kantorovich problem}
Relaxing the map constraint, the Kantorovich formulation seeks probability measures $P$ on $\Omega^* \times \Omega$ that attains the infimum,
\begin{equation}
\label{eq: KP}
\text{(KP)} \quad \inf \{ \int_{\Omega^* \times \Omega} c(\xi, x) dP(\xi, x) \mid P\in \Gamma(\lambda, \mu) \} .
\end{equation}
where $ \Gamma (\lambda, \mu )$ denotes the collection of all probability measures on $\Omega^* \times \Omega $ with marginals $\lambda$ on $\Omega^*$ and $\mu$ on $\Omega$, and \( c: \Omega^* \times \Omega \rightarrow \mathbb{R}^+ \) is transportation cost function.
The existence and uniqueness are guaranteed with similar assumptions (c.f. Theorem 1.17 \cite{santambrogio2015optimal}).

In the discrete setting (using the same notation \( \Xi, \mathcal{X}, a, b \) as in the Monge formulation), the cost function is evaluated as a matrix \(M \in \mathbb{R}^{n_1 \times n_2}\), with entries
\begin{equation}
M_{ij} = c(\xi_i, x_j), \quad 1 \le i \le n_1, \, 1 \le j \le n_2.    
\end{equation}
The Kantorovich problem then reduces to the following linear program:
\begin{align}
\label{wasserstein}
\text{(KP)} \quad  
\min_{P \in \Gamma(a,b)} \langle P, M \rangle \notag = \min_{P \in \Gamma(a,b)} \sum_{i=1}^{n_1} \sum_{j=1}^{n_2} M_{ij} \cdot P_{ij} \notag 
\end{align}
where \(\Gamma(a, b)\) represents the set of all feasible coupling matrices, defined as the discrete probability measures \(\Gamma(a, b) = \{ P \geq 0 \mid P\mathbf{1}_{n_2} = a, P^T\mathbf{1}_{n_1} = b \}\). Moreover, the following sparity property holds for the discrete implementation of the Kantorovich formulation \citep{peyre2019computational}:

\begin{lemma}[Sparsity of Transport Plan]
The solution to the linear programming is sparse; the number of non-zero entries in the transport plan \( P \) is at most \( n_1 + n_2 - 1 \).
\end{lemma}

In practice, optimal transport plan is often computed with entropy regularization using the Sinkhorn Algorithm \citep{cuturi2013sinkhorn}. In this case, the smoothened transport plan is not sparse anymore. 
\section{Geometry Embedding as Optimal Transport}

In this section, our goal is to embed a complex geometric domain into a simpler latent geometric domain while simultaneously embedding the associated density function. To address this challenge, we construct a computational framework based on optimal transport, as described below. 

For convenience, we use \( f \), $u$ and $\Omega$ to denote the $f^{\text{sub}}$, $u^{\text{sub}}$ and $\partial \Omega_f$ in Eq~\eqref{eq:sub-manifold problem}. According to the settings from Sec~\ref{problem setting}, $f$ is a density function defined on the complex geometric domain \( \Omega \) and $\Omega = \operatorname{supp}(f)$. We define a measure \( \mu \) built from the density \( f \) as follows:
\begin{equation}
    d\mu = f(x) \, dx \quad \text{on } \Omega.
\end{equation}
The task is then to embed the physical density function \( f \) on $\Omega$ into a latent density function $g$ on a simple geometric domain $\Omega^*$ within the same metric space $\mathbb{R}^d$. This is equivalent to finding a transformation between measures $\mu$ and $\lambda$, where \( d\lambda=g(\xi)d\xi \) represents a uniform measure on a canonical geometric domain \( \Omega^* \), such as a unit sphere or torus. And we use $x$ and $\xi$ to represent positions in $\Omega$ and $\Omega^*$.



Thereby, we can model the geometries as the density functions (probability measures) and then encode these density functions using transport maps/plans, finally cooperate with the latent operators such as FNO to solve the PDEs, detailed in the following subsection.

\subsection{Methodological Formulations}
Instead of learning the map directly from the measure $\mu$ to the solution functions $u$, we encode $\mu$ to the corresponding optimal transport $T: \lambda \to \mu$, where $\lambda$ is the reference uniform measure. $T$ can be viewed as the reparameterization of $\mu$.

\paragraph{Transport Map}

Given a transport map $T$ from latent measure $\lambda$ to physical measure $\mu$, which is a function defined on latent domain $\Omega^*$ :
\begin{equation}\label{eq: transport map}
    \begin{aligned}
        T: (\Omega^*, \lambda) &\to (\Omega, \mu),\\
        \xi & \mapsto x,
    \end{aligned}
\end{equation}
the encoder is defined as the optimal transport solver that maps the density functions  to the associated transport maps
\begin{equation*}
\begin{split}
    \mathcal{Q}: L^1(\Omega; \mathbb{R}) &\to L^1(\Omega^*; \mathbb{R}^d)\\
    f &\mapsto T.
\end{split}
\end{equation*}
Note, in particular, that uniqueness of the optimal transport map makes this mapping well-defined. The latent neural operator is then defined such that it maps the transport map $T$ to a latent solution function $v$ on the latent domain $\Omega^*$:
\begin{equation}\label{latent operator - map}
    \begin{aligned}
        \mathcal{G}^*: T & \mapsto v \quad \text{on } \Omega^*.
    \end{aligned}
\end{equation}
We then define the decoder $\mathcal{P}: v \mapsto u$ as 
\begin{equation}\label{solutoin-map}
    u(x) = v \circ T^{-1}(x) \quad \forall x\in \Omega.
\end{equation}
The composition $\mathcal{P} \circ \mathcal{G}^* \circ \mathcal{Q} : f \mapsto u$ defines our approximation to the operator $\mathcal{G}^\dagger$ from equations \eqref{eq:manifold-problem} and \eqref{eq:sub-manifold problem}.

Moreover, for any function \( a(x) \) on the physical domain \( \Omega \), the transport map \( T \) enables encoding it onto the latent domain \( \Omega^* \) through:
\begin{equation}\label{normal-map}
    a^*(\xi) = a(T(\xi)), \quad \forall \xi \in \Omega^*.
\end{equation}

\paragraph{Transport Plan}

 

Given a transport plan $P$ from latent measure $\lambda$ to physical measure $\mu$, which is a probability measure on $\Omega^* \times \Omega$ with marginals $\mu$ on $\Omega$ and $\lambda$ on $\Omega^*$:
\begin{equation}
    \begin{aligned}
        P: (\Omega^*, \lambda)\times(\Omega, \mu) &\to [0,1],\\
        (\xi,x) & \mapsto P(\xi,x),
    \end{aligned}
\end{equation}
the encoder is defined as the optimal transport solver that maps the density functions  to the marginal of the transport plans
\begin{equation*}
\begin{split}
    \mathcal{Q}: L^1(\Omega; \mathbb{R}) &\to L^1 (\Omega^*; \mathbb{R}^d)\\
    f &\mapsto \int_{\Omega} P (\cdot,x) x d\mu(x).
\end{split}
\end{equation*}
The latent neural operator is defined such that maps the marginal map of $P$ to the latent solution function $v$ on the latent domain $\Omega^*$:
\begin{equation}\label{latent operator - plan}
    \begin{aligned}
        \mathcal{G}^*: \int_{\Omega} P (\cdot,x) x d\mu(x) & \mapsto v \quad \text{on } \Omega^*,
    \end{aligned}
\end{equation}
noting that \( \int_{\Omega} P(\cdot, x)\, x\, d\mu(x) \) is a function mapping from the latent domain to the physical domain. It serves a similar role to the transport map \( T \) in Eq.~\eqref{eq: transport map}, transporting points from the latent space to the physical space.
The decoder $\mathcal{P}: v \mapsto u$ is defined as 
\begin{equation}\label{solution-map}
    u(x) = \int_{\Omega^*} P(\xi,x)v(\xi)  d\lambda(\xi) \quad \forall x\in \Omega .
\end{equation}
As before, we define our approximate operator as $\mathcal{P} \circ \mathcal{G}^* \circ \mathcal{Q} : f \mapsto u$.

Moreover, for any function \( a(x) \) on the physical domain \( \Omega \), we note that the transport plan \( P \) enables encoding it onto the latent domain \( \Omega^* \) as
\begin{equation}\label{normal-plan}
    a^*(\xi) = \int_{\Omega} P (\xi,x) a(x) d\mu(x).
\end{equation}

So far, we have formulated an approximation space of solution operators as a combination of two components: learning the transport map/plan, and learning the latent neural operator. In application, we adopt numerical OT methods to learn the transport map or plan, using the squared Euclidean distance as the transportation cost function.

\paragraph{Transportation Cost function}
We choose the squared Euclidean distance as the cost function in optimal transport due to both its mathematical convenience and its relevance in geometric applications. In the context of transporting probability measures between two geometric domains embedded in 3D space, the squared Euclidean distance \( c(x, y) = \|x - y\|^2 \) naturally captures the geometric cost of moving mass from one location to another. Moreover, it is widely used in the literature and aligns with the assumptions of classical results like Brenier's theorem, which guarantees the existence of a unique optimal map under this cost. Thus, it serves as a principled and standard choice for geometric transformation tasks in 3D Euclid space.

\subsection{Method Generalization}
Interestingly, state-of-the-art methods like Geo-FNO and GINO can also be viewed as special cases within our proposed framework. In the following subsections, we categorize them into two types—\textit{map-type} and \textit{plan-type}—which align with the above two formulations, respectively.

\paragraph{Map-type: Geo-FNO}


As for Geo-FNO, the deformation map can be viewed as generalized OT map with a special choice of cost function:
\begin{equation}
    \int_{\Omega} c(x, T^{-1}(x)) d\mu(x) := \int_{\Omega} G(T^{-1}(x)) - u^{\text{sub}}(x) d\mu(x)
\end{equation}
where $u^{\text{sub}}$ is our target solution function as presented in Eq~\eqref{eq:sub-manifold problem}. 
However, this is a non-standard, generalized cost function, since it depends not on $(x, T^{-1}(x))$, but the global solution operator $G$ and domain $\Omega$.

Similar to learning the function \(\phi\) (defined in Theorem~\ref{thm:brenier}) in  Monge's formulation, Geo-FNO  implements a skip connection \(\xi = \psi(x) + x\) within the deformation network to learn only \(\psi\). The difference is that Geo-FNO adopts an end-to-end approach, optimizing the deformation map based on the final solution error and learning a shared network to implement geometry deformation. In contrast, the Monge problem learns the deformation map separately from the operator learning and optimizing based on the instance-specific transportation cost for each geometry respectively. Despite these differences in optimization strategies, both approaches can be categorized as employing a \textit{map-type }for the deformation layer.



\paragraph{Plan-type: GINO}

Similar to learning the optimal coupling matrix in the Kantorovich formulation, GINO employs a Graph Neural Operator (GNO), which can be interpreted as a learnable Graph Laplacian. GNO constructs the adjacency matrix by performing neighborhood searches within a fixed radius and learns the edge weights using kernel functions. In contrast, the Kantorovich formulation solves a global transport plan (optimal coupling matrix) by optimizing the Wasserstein distance, which accounts for the pairwise distances between all points across the two domains. Despite these differences in optimization strategies, both approaches can be categorized as employing a \textit{plan-type} transformation layer.

\section{Optimal Transport Neural Operator (OTNO)}
\label{otno}
In this paper, we introduce a novel model, the optimal transport neural operator (OTNO), which efficiently integrates optimal transport with neural operators. Our model employs the Projection pursuit Monge map to obtain an approximate solution of the Monge OT map $T$, and employs Sinkhorn method \citep{cuturi2013sinkhorn} to obtain an approximate solution of the Kantorovich OT plan \(T\). The resulting Map/plan is then utilized to construct the OT encoder/decoder for neural operator. The methodology and implementation are detailed below.

\subsection{OTNO Algorithm}
Given a dataset $\{(\mathcal{X}_j, u_j)\}_{j=1}^N$ of surface sampling meshes and corresponding PDEs' solution values on surface, where $\mathcal{X}_j = \left[ x_{j,k} \right]_{k=1}^{n_j^1} \in \mathbb{R}^{n_j^1\times 3}$ and $u_j = \left[ u_{j,k} \right]_{k=1}^{n_j^1} \in \mathbb{R}^{n_j^1\times s}$. For each $\mathcal{X}_j$, generate a latent surface mesh $\Xi_j=\left[ \xi_{j,l} \right]_{l=1}^{n_j^2} \in \mathbb{R}^{n_j^2\times 3}$ using a 2D parametric mapping from a square grid, where $n_j^2$ is a perfect square. Compute the normals $\mathcal{N}_j \in \mathbb{R}^{n_j^1\times 3}$ on $\mathcal{X}_j$ and $\mathcal{H}_j \in \mathbb{R}^{n_j^2\times 3} $ on $\Xi_j$. 

By solving OT problem from each latent mesh $\Xi_j$ to each physical sampling mesh $\mathcal{X}_j$, we obtain a set of transported mesh $\{\mathcal{X}_j'\}_{j=1}^N$, where $\mathcal{X}_j' =\left[ x'_{j,l} \right]_{l=1}^{n_j^2} \in \mathbb{R}^{n_j^2\times 3} $. Note that $\mathcal{X}_j'$ is the transported mesh obtained by applying the OT map/plan to the latent mesh. The transported mesh serves as an representation of the physical surface from which the mesh $X_j$ is sampled.

The following algorithm is detailed in Algorithm~\ref{algorithm:OTNO}, applicable for both OT map and OT plan scenarios. The primary distinctions between utilizing an OT map and an OT plan are in the generation of the latent mesh $\Xi_j$ and the construction of the transported mesh $\mathcal{X}_j'$. These distinctions are elaborated in Sections~\ref{otno-plan} and~\ref{otno-map}. This section assumes the availability of these meshes and focuses on illustrating the overarching algorithm.

\begin{algorithm}[h]
\caption{Optimal Transport Neural Operator (OTNO) }
\label{algorithm:OTNO}
\begin{algorithmic}[1]
\State Given physical mesh $\{\mathcal{X}_j \}_{j=1}^N$, latent mesh $\{\Xi_j\}_{j=1}^N$, transported mesh $\{\mathcal{X}_j'\}_{j=1}^N$ and solution values $\{u_j\}_{j=1}^N$.
\State Initialize a FNO $\mathcal{G}_\theta$. 
\For{$j = 1$ to $N$}
    \State \textbf{1. Build Index Mapping:}
    \State \quad Encoder indices $\mathcal{E} = \left( \arg \underset{k=1,\dots, n_j^1}{\text{min}} \| x'_{j,l}-x_{j,k} \|_2: l=1,\dots, n_j^2 \right)$
    \State \quad Decoder indices $\mathcal{D} = \left( \arg \underset{l=1,\dots, n_j^2}{\text{min}} \| x'_{j,l} - x_{j,k} \|_2: k=1,\dots, n_j^1 \right)$
    
    \State \textbf{2. OT encoder}: $\mathcal{M}_j = \mathcal{X}_j(\mathcal{E}) \in \mathbb{R}^{n_j^2\times 3}$, where \(\mathcal{M}_j\) selects rows from \(\mathcal{X}_j\) according 
    \phantom{\quad \quad \textbf{2. OT encoder}:} to the indices specified in $\mathcal{E}$.
    \State \textbf{3. Latent FNO}: $v_j = \mathcal{G}_\theta(\mathcal{T}_j)$, where $\mathcal{T}_j  = (\Xi_j, \mathcal{M}_j, \mathcal{H}_j \times \mathcal{N}_j (\mathcal{E}) ) \in \mathbb{R}^{n_j^2\times 9}$ \\
    \phantom{\quad \quad \textbf{3. Latent FNO}:} ($ \mathcal{H}_j \times \mathcal{N}_j (\mathcal{E})$ computes the cross product between rows).
    \State \textbf{4. OT decoder}: $u_j' = v_j(\mathcal{D}) \in \mathbb{R}^{n_j^1\times s}$, where \(u_j\) selects rows from \(v_j\) according 
    \phantom{\quad \quad \textbf{2. OT encoder}:} to the indices specified in $\mathcal{D}$.
\EndFor
\State Compute the empirical loss over all dataset instances: $\displaystyle \sum_{j=1}^N \|u_j' - u_j \|_{\mathcal{U}} $. 
\end{algorithmic}
\end{algorithm}

\paragraph{Encoder}
For each point $x'_{j,l}$ ($l=1,\dots,n_j^2$) in the transported mesh $\mathcal{X}_j'$, we find the closet point in $\mathcal{X}_j$ and denote $e_k$ as the index of this closest point in $\mathcal{X}_j$. Thus, we obtain an index sequence
$\mathcal{E} = \left( \arg \underset{k=1,\dots, n_j^1}{\text{min}} \| x'_{j,l}-x_{j,k} \|: l=1,\dots, n_j^2 \right) = (e_1,\dots,e_{n_j^2} ).$
Using these indices, we encode the mesh \(\mathcal{X}_j\) to \(\mathcal{M}_j = \mathcal{X}_j(\mathcal{E})\in \mathbb{R}^{n_j^2\times 3}\), where \(\mathcal{M}_j\) selects rows from \(\mathcal{X}_j\) according to the indices specified in $\mathcal{E}$. 
\paragraph{Latent Operator}
In the latent space, we deploy the FNO $\mathcal{G}_{\theta}$ to execute the latent neural operator $\mathcal{G}$ as introduced in Eq~\eqref{latent operator - map}\eqref{latent operator - plan}. Denote $T$ as the transport map in Eq~\eqref{latent operator - map} or marginal map of transport plan in Eq~\eqref{latent operator - plan}. Then it can be fundamentally represented by pairs of latent mesh and transported mesh $(\Xi_j, T(\Xi_j)) = (\Xi_j, \mathcal{X}_j')$. Considering $T$ as a deformation map, we include deformation-related features by using the cross-product over normals $\mathcal{H}_j \times T(\mathcal{H}_j) $ (further discussion on normal feature is in Sec~\ref{normal features}). Thus, we configure $\mathcal{T}_j$ as $(\Xi_j, \mathcal{X}_j', \mathcal{H}_j \times T(\mathcal{H}_j ))$ to fully encapsulate the map's properties. To enhance the quantity of surface representation, we substitute points in $\mathcal{X}'_j$ with their closest counterparts in $\mathcal{X}_j$, i.e. use $\mathcal{M}_j = \mathcal{X}_j(\mathcal{E})$ to replace $\mathcal{X}_j'$. Similarly, we use $\mathcal{N}_j (\mathcal{E})$ to replace the $T(\mathcal{H}_j )$. Consequently, $\mathcal{T}_j = (\Xi_j, \mathcal{M}_j, \mathcal{H}_j \times \mathcal{N}_j (\mathcal{E}) )$ serves as a comprehensive representation of the deformation map $T$, as well as the input for the latent FNO.

\paragraph{Decoder}
Corresponding to the encoder, we compute the index $d_k$ of the closest point in the transported mesh $\mathcal{X}_j'$ for each point $x_{j,k}$ ($k=1,\dots,n_j^1$) in $\mathcal{X}_j$ and build a decoder index sequence $\mathcal{D} = \left( \arg \underset{l=1,\dots, n_j^2}{\text{min}} \| x'_{j,l} - x_{j,k} \|_2: k=1,\dots, n_j^1 \right) = (d_1,\dots,d_{n_j^1} ) $. Using these indices, we decode the solutions $v_j$ back to physical surface.

Note that the latent surface mesh $\Xi_j \in \mathbb{R}^{n_j^2 \times 3}$ is constructed from a 2D parametric grid. This ensures that the input features $T_j \in \mathbb{R}^{n_j^2 \times 9}$ for the FNO are organized on a 2D grid. Consequently, the FNO computations occur in 2D space, leveraging the structured layout of $\Xi_j$, rather than directly handling the unstructured 3D point cloud.

\subsection{OTNO - Kantorovich Plan}\label{otno-plan}
\subsubsection{Transported Mesh}
Using the Sinkhorn method \citep{cuturi2013sinkhorn} detailed in the following Sec~\ref{sinkhorn}, we obtain dense coupling matrices \(P_j\) that represent the OT plans from the latent computational mesh \(\Xi_j\) to the boundary sampling mesh \(\mathcal{X}_j\). How to effectively utilize these dense matrices within a neural operator framework become a problem worth discussing.

As described in Eq.~\eqref{latent operator - plan}, the transported mesh is obtained by applying the function \( \int_{\Omega} P(\cdot, x)\, x\, d\mu(x) \) to the latent mesh. Therefore, in the discrete case, the transported mesh is given by \( \mathcal{X}_j' = P_j \mathcal{X}_j \). However, directly applying the dense matrix by multiplying it with the mesh will lead to a lower accuracy of the final predictions because it only approximates the solution plan for the Kantorovich problem \eqref{eq: KP}. And saving large-scale dense matrices is costly. Therefore we discuss how to use these optimal coupling matrices more effectively. As the matrix \(P_j\) represents a discrete probability measure on \(\mathcal{X}_j \times \Xi_j\), a practical approach is to focus on the maximum probability elements, referred to as the "Max" strategy. Additionally, a more effective "Mean" strategy involves replacing each point \(x' \in \mathcal{X}_j'\) with the nearest point in \(\mathcal{X}_j\), significantly reducing approximation ambiguity caused by Sinkhorn method.  Further, we can encode/decode by the top-k nearest neighbors in $\mathcal{X}_j$/$\mathcal{X}'_j$ instead of only find the single nearest one. Details of these different strategies are further discussed in Section~\ref{ot_strategy}. 


\subsubsection{Latent Mesh}
\begin{figure}[h]
    \centering
    \includegraphics[width=1\linewidth]{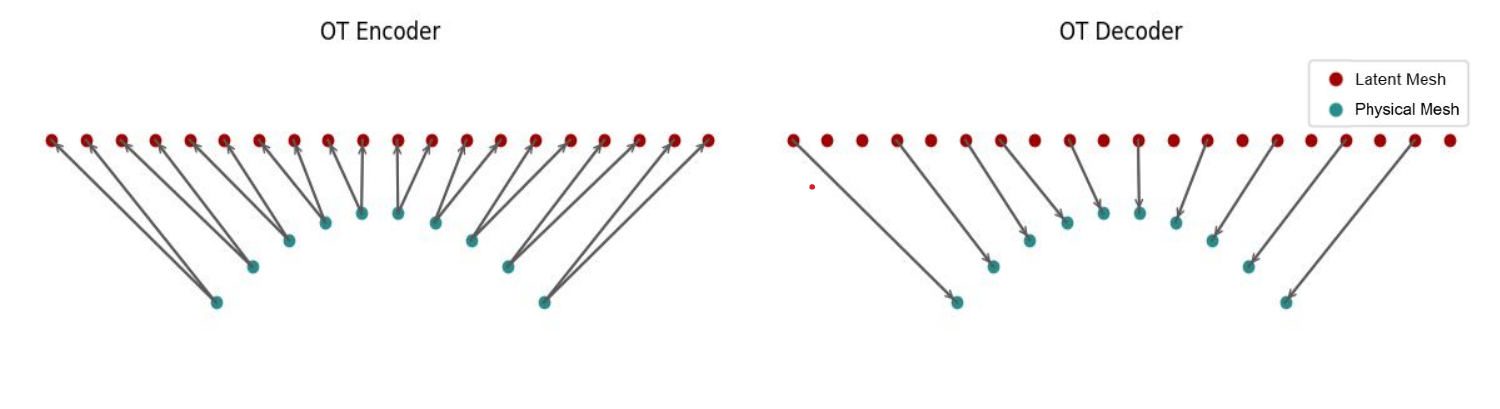}
    \caption{A simple illustration for OT encoder and OT decoder. The curve represents the boundary sampling points and the line denotes the latent computational grid}
    \label{fig:ot_encoder_decoder}
\end{figure}

In practice, we design the computational grid to have more vertices than the boundary sampling mesh to ensure maximal information retention during the encoding and decoding processes. Let \(\alpha\) be an expansion factor, and the number of points in latent space is then set to $n_j^2 = \lceil \sqrt{\alpha \times n_j^1} \rceil^2$. This approach implies that the encoder functions as an interpolator, while the decoder acts as a selective querier. A simple illustration of the encoder and decoder processes is shown in Fig.\ref{fig:ot_encoder_decoder}. And detailed ablation studies for latent mesh shape and latent mesh size are in Sec~\ref{latent shape} and Sec~\ref{latent size}.


\subsubsection{Sinkhorn Algorithm}
\label{sinkhorn}
Sinkhorn \citep{cuturi2013sinkhorn} method added an entropy regularizer to the Kantorovich potential and greatly improved efficiency. They first propose the Sinkhorn distance that added entropy constraint:
\begin{equation}
    d_{M,\alpha} (a,b) = \min_{P \in \Gamma_{\alpha}(a,b)} \langle P, M \rangle ,
\end{equation}
where
\begin{equation}
    \Gamma_{\alpha}(a,b)= \{ P \in \Gamma (a,b) \mid \textbf{KL}(P,ab^T) \leq \alpha \} = \{ P \in \Gamma (a,b) \mid h(P) \geq h(a) + h(b) - \alpha \} .
\end{equation}
Then they consider the dual problem that arises from Lagrange multiplier:
\begin{equation}
    P^\beta= \underset{P\in \Gamma(a,b)}{\text{argmin}} \quad  \langle P, M \rangle - \frac{1}{\beta} h(P),
\end{equation}
This formulation leads to a problem where \( \beta \) adjusts the trade-off between the transport cost and the entropy of the transport plan \( P \). When \( \beta \) increases, the influence of the entropy regularization decreases, making \( P^\beta \) converge closer to the solution of the original Kantorovich problem \eqref{eq: KP}. This implies that a larger \( \beta \) leads to a solution that is more accurate and economically efficient.

By introducing the entropy constraint, the Sinkhorn distance not only regularizes the OT problem but also ensures that the solution is computationally feasible even for large-scale problems. This regularization dramatically improves the numerical stability and convergence speed of the algorithm.

\paragraph{Implementation}
In the implementation, we set \( a = \frac{1}{n_1} \mathbf{1}_{n_1} \), assuming each point in the latent space contributes equally to computations. We set \( b = \frac{1}{n_2} \mathbf{1}_{n_2} \), as the sampling mesh \( \mathcal{X} = \{x_1, \dots, x_{n_1}\} \), obtained from CFD simulations, typically offers increased point density in regions with sharp variations. This uniform mass vector on \( \mathcal{X} \) ensures a denser representation in these critical areas, improving the accuracy of aerodynamic predictions. For cases with excessively dense regions, we employ voxel downsampling to prevent excessive density variations, thereby maintaining the feasibility of our uniform mass vector. We set \( \beta = 1 \times 10^6 \) to ensure that the solution is economical without incurring excessive computational costs. 
For computational support, we utilize the geomloss implementation from the Python Optimal Transport (POT) library \citep{flamary2021pot}, which supports GPU acceleration and use lazy tensors to store dense coupling matrix obtained from Sinkhorn method.

\paragraph{Voxel Downsampling}  
To achieve a well-balanced input density function, we employ voxel downsampling as a normalization process that constrains the density function within a predefined range $[0, a]$, where $a$ is determined by the voxel size. This approach mitigates excessive clustering in regions with high point density, such as around the wheels in some some car data, therefore ensuring a more uniform spatial distribution of points.

\subsection{OTNO - Monge Map}\label{otno-map}
\subsubsection{Transported Mesh}
Using the PPMM, detailed in Section~\ref{ppmm}, we obtain the transported mesh \( X_j' \), which is mapped from the latent mesh \( \Xi_j \) to the physical space. This results in the same number of points as the latent mesh while representing the distribution of the physical mesh. Since PPMM operates directly on the latent mesh, projecting it towards the physical mesh (as shown in Algorithm~\ref{algorithm:ppmm}), the output is the transported mesh itself rather than a mapping that can operate on any function on the latent mesh, and it cannot offer an inverse direction itself. (An OT plan, discretized as a coupling matrix, can handle both.) Therefore, we also compute the indices for encoding and decoding, as outlined in Algorithm~\ref{algorithm:OTNO}.

\subsubsection{Latent Mesh}
Since the latent mesh has the same number of points as the physical mesh, and the 2-dimensional FNO on the latent space requires a square mesh, the number of points in both the latent mesh and the physical mesh should be a perfect square.

Therefore, we first apply voxel downsampling to normalize the mesh density, followed by random sampling to further downsample the mesh so that ensure the number of points is a perfect square. Finally, we generate the latent mesh to match the square number of points in the downsampled physical mesh. We choose a spherical mesh as the latent mesh, given its suitability for the Projection Pursuit Monge Map (PPMM).

\subsubsection{Projection pursuit Monge map (PPMM)}
\label{ppmm}

\begin{algorithm}
\caption{Projection pursuit Monge map}
\label{algorithm:ppmm}
\begin{algorithmic}
\State \textbf{Input:} two matrix $X \in \mathbb{R}^{n \times d}$ and $Y \in \mathbb{R}^{n \times d}$
\State $k \leftarrow 0$, $X_0 \leftarrow X$
\Repeat
    \State (a) calculate the most 'informative' projection direction $e_k \in \mathbb{R}^d$ between $X_k$ and $Y$ 
    \State (b) find the one-dimensional OT Map $\phi^{(k)}$ that matches $X_ke_k$ to $Ye_k$ (using look-up table)
    \State (c) $X_{k+1} \leftarrow X_k + (\phi^{(k)}(X_ke_k) - X_ke_k^T)e_k^T$ and $k \leftarrow k + 1$
\Until{converge}
\State The final mapping is given by $\hat{\phi}: X \rightarrow X_k$
\end{algorithmic}
\end{algorithm}
The PPMM proposes an estimation method for large-scale OT maps by combining the concepts of projection pursuit regression and sufficient dimension reduction. As summarized in Algorithm~\ref{algorithm:ppmm}, in each iteration, the PPMM applies a one-dimensional OT map along the most "informative" projection direction. The direction \(e_k\) is considered the most "informative" in the sense that the projected samples \(X_k e_k\) and \(Y e_k\) exhibit the greatest "discrepancy." The specific method for calculating this direction is detailed in Algorithm~1 in paper \cite{meng2019large}.
\section{Experiments}
We conducted experiments on three CFD datasets. Two of them are 3D car datasets, where the target prediction is the pressure field or drag coefficient, which depends solely on the car surface—a 2D manifold. The \textbf{ShapeNet} dataset includes 611 car designs, each with 3.7k vertices and corresponding average pressure values, following the setup from \cite{GINO}. The \textbf{DrivAerNet} dataset, sourced from \cite{elrefaie2024drivaernetparametriccardataset}, contains 4k meshes, each with 200k vertices, along with results from CFD simulations that measure the drag coefficient. Additionally, we further evaluate our model on the 2D \textbf{FlowBench} dataset \citep{tali2024flowBench}, which features a wider variety of shapes, including three groups, each containing 1k shapes with a resolution of 
$512 \times 512$. Although the PDE solutions are not on the boundary sub-manifold, preventing our model from reducing dimensions by embedding boundary geometries, this dataset provides an excellent opportunity to explore our model's capabilities across diverse shapes. The code for these experiments is available at \url{https://github.com/Xinyi-Li-4869/OTNO}.

\subsection{Main Experiments - 3D Car Datasets}
\subsubsection{ShapeNet Car Dataset}
\begin{figure}[t]
    \centering
    \begin{minipage}{1.0\textwidth}
        \centering
        \includegraphics[width=\linewidth]{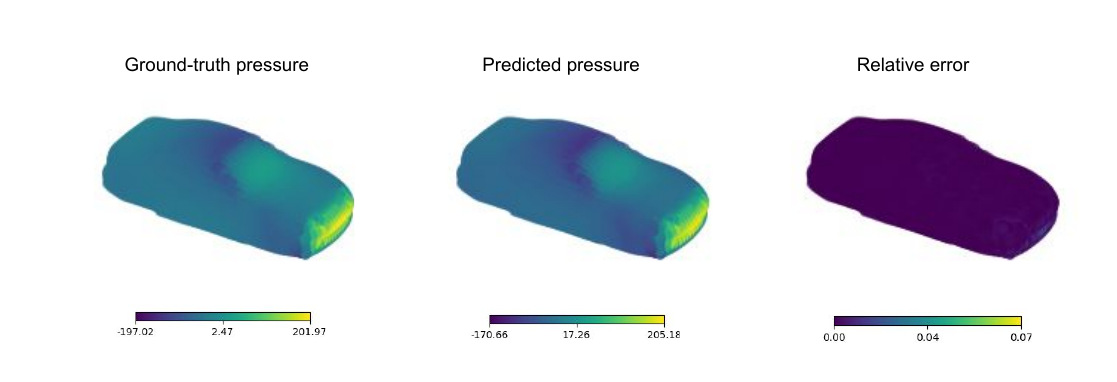}
        \subcaption{Results of pressure on ShapeNet Car dataset.}
        \label{fig:visual_shape}
    \end{minipage}\\
    \begin{minipage}{0.6\textwidth}
        \centering
        \includegraphics[width=\linewidth]{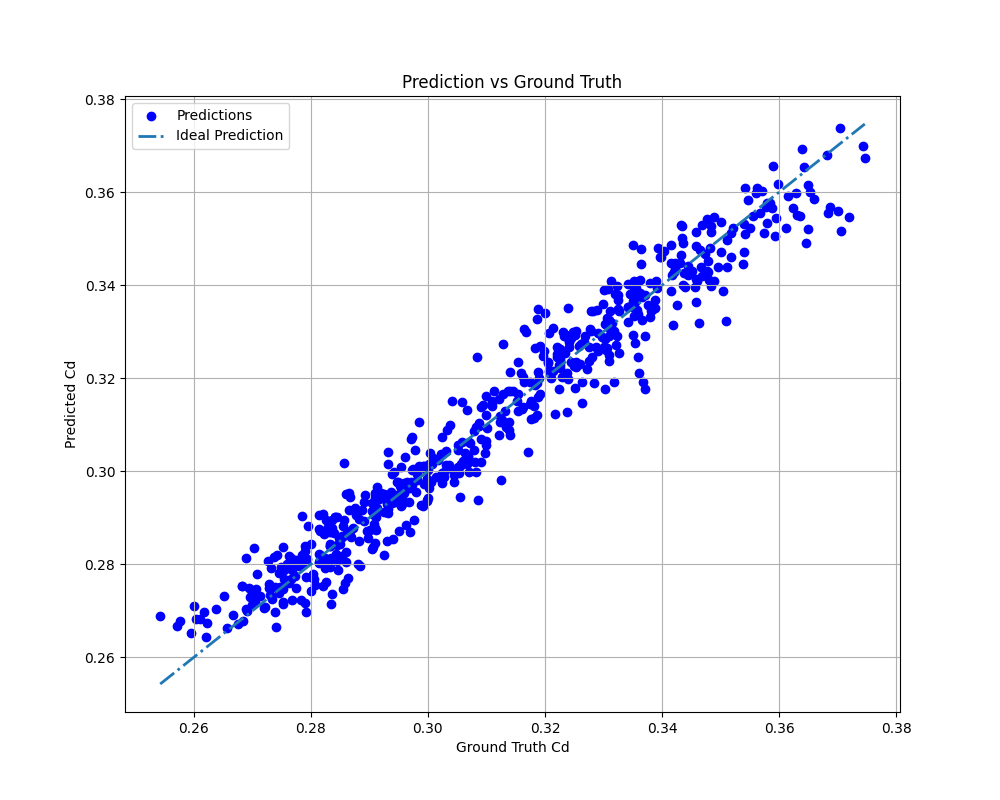}
        \subcaption{Results of drag coefficient on DrivAerNet Car dataset}
        \label{fig:visual_drivaer}
    \end{minipage}\\
    \caption{Results visualization for OTNO on Car Datasets}
    \label{fig:result_visualization}
\end{figure}

To ensure a fair comparison, we maintained identical experimental settings to those used in GINO paper. We compared the relative error, total time (including data processing and training), and GPU memory usage against key baselines from \cite{GINO}. As detailed in Table.\ref{table:shape}, our method, OTNO, achieved a relative error of \textbf{6.70\%}, which is a slight improvement over the lowest error reported in the baseline—\textbf{7.21\%} by GINO. Across all baselines, our method demonstrated reduced total time and lower GPU memory usage. Compared to GINO, our method significantly reduced both time costs and memory expenses by factors of eight and seven, respectively. The visual results for pressure prediction are presented in Fig.\ref{fig:visual_shape}.
\begin{table}[h]
  \caption{Predict pressure field on ShapeNet Car Dataset (single P100)}
  \label{table:shape}
  \centering
  \begin{tabular}{lccc} 
    \toprule
    Model    & Relative L2   & Total Time (hr) & GPU Memory (MB)\\
    \midrule
    Geo-FNO   & 13.50\%  & 1.96     &  12668 \\
    UNet   & 13.14\%   & 6.86   &  7402\\
    GINO     & 7.21\%    & 10.45    & 12734 \\
    OTNO(Plan)   & \textbf{6.70\%}    & \textbf{1.52}    & \textbf{1890} \\
    
    \bottomrule
  \end{tabular}
\end{table}

\subsubsection{DrivAerNet Car Dataset}
\label{drivaernet dataset}
Herein, we follow all the experimental settings from DrivAerNet paper \citep{elrefaie2024drivaernetparametriccardataset} and compare our model to RegDGCNN, as proposed in \cite{elrefaie2024drivaernetparametriccardataset}, and the state-of-the-art neural operator model, GINO. Note that we do not use pressure data for training; instead, we only use the drag coefficient (Cd), with further details provided in Appendix~\ref{appendix: Cd}. The visual result for Cd prediction is presented in Fig.\ref{fig:visual_drivaer}. For OTNO, we employ voxel downsampling with a size of 0.05, corresponding to approximately 60k samples (see ablation in \ref{ablation: physical sampling}). 

As detailed in Table.\ref{table:drivaer}, OTNO(Plan) demonstrates significantly superior accuracy and reduced computational costs compared to RegDGCNN, halving the MSE, speeding up computations by a factor of 5, and reducing GPU memory usage by a factor of 24. Compared to GINO, OTNO(Plan) achieves a slightly lower MSE, a marginally higher R2 score, and notably reduces total computation time and memory usage by factors of 4 and 5, respectively.

Given that PPMM is designed for large-scale OT maps, we evaluate OTNO(Map) on this large-scale dataset to assess its performance. Unfortunately, both the error and time cost are worse than OTNO(Plan). However, the memory cost is significantly lower. This is primarily because OTNO(Plan) expands the latent space, while OTNO(Map) does not.

\begin{table}[h]
  \centering
  \caption{Predict drag coefficient (Cd) on DrivAerNet Car Dataset (single A100). The results for the first six baselines are reproduced from \cite{choy2025factorized}.}
  \label{table:drivaer}
  \begin{tabular}{lcccc} 
    \toprule 
    \scriptsize{\textbf{Model}}	
    &\scriptsize{\textbf{MSE (e-05)}}	
    &\scriptsize{\textbf{R2 Score}}  
    &\scriptsize{\textbf{Total Time (hr)}}   
    & \scriptsize{\textbf{Memory (MB)}}\\
    \midrule
    \scriptsize{\textbf{PointNet++\citep{qi2017pointnet++}}} & \small{7.81} & \small{0.896} & \small{-} & \small{-} \\
    \scriptsize{\textbf{DeepGCN\citep{li2019deepgcns}}} & \small{6.30} & \small{0.916} & \small{-} & \small{-} \\
    \scriptsize{\textbf{MeshGraphNet\citep{pfaff2020learning}}} & \small{6.00} & \small{0.917} & \small{-} & \small{-} \\
    \scriptsize{\textbf{AssaNet\citep{qian2021assanet}}} & \small{5.43} & \small{0.927} & \small{-} & \small{-} \\
    \scriptsize{\textbf{PointNeXt\citep{qian2022pointnext}}} & \small{4.58} & \small{0.939} & \small{-} & \small{-} \\
    \scriptsize{\textbf{PointBERT\citep{yu2022point}}} & \small{6.33} & \small{0.915} & \small{-} & \small{-} \\
    \scriptsize{\textbf{RegDGCNN\citep{elrefaie2024drivaernetparametriccardataset}}} & \small{6.63} & \small{0.887} & \small{10.78} & \small{72392} \\
    \scriptsize{\textbf{GINO\citep{GINO}}} & \small{3.33} & \small{0.955} & \small{7.73} & \small{14696} \\
    \scriptsize{\textbf{OTNO(Map)}} & \small{3.93} & \small{0.947} & \small{10.63} & \small{\textbf{2896}} \\
    \scriptsize{\textbf{OTNO(Plan)}} & \small{\textbf{3.28}} & \small{\textbf{0.956}} & \small{\textbf{5.26}} & \small{9702} \\
    \bottomrule
  \end{tabular}
\end{table}

\subsection{Showcase of Dataset with Diverse Geometries - 2D Flow Datasets}
To assess the performance of instance-dependent deformation in our model across diverse geometries, we conducted further evaluations using the FlowBench dataset \citep{tali2024flowBench}, The experimental settings are in Appendix~\ref{appendix: flowbench}. We use two metrics for training:

\begin{enumerate}
    \item M1: \textit{Global metrics}: The errors of in velocity and pressure fields over the entire domain.
    \item M2: \textit{Boundary layer metrics}: The errors in velocity and pressure fields over a narrow region around the object. We define the boundary layer by considering the solution conditioned on the Signed Distance Field ($0 \leq SDF \leq 0.2$). 
\end{enumerate}

The results of training using the M1 metric (global) and M2 metric (boundary) are presented in Table~\ref{table:flowbench-m1}, and Table~\ref{table:flowbench-m2}, respectively. We present a subset of the visualization results in Fig.~\ref{fig:flowbench_visualization}, with the full set of prediction plots provided in Appendix\ref{appendix: flowbench}.

The results demonstrate that OTNO significantly outperforms in accuracy under both the M1 (global) and M2 (boundary) metrics, particularly for M1. Furthermore. And OTNO achieves notably better accuracy across all three groups, especially for G2 (harmonics). However, cost reduction is not observed in the FlowBench dataset. It is important to note that the cost reductions seen in car datasets stem from the sub-manifold method, which employs optimal transport to generate 2D representations and perform computations in 2D latent space instead of 3D. In the FlowBench dataset, however, the solutions are not restricted to the boundary sub-manifold. Even for the M2 metric, although the relevant data is close to the boundary with a width of 0.2, it does not reduce to a 1D line. As a result, computations cannot be confined to the sub-manifold, and the associated cost reduction benefits are consequently absent.

We do not present Geo-FNO results on this dataset as the relative L2 errors consistently exceed 60\%. Our analysis suggests that Geo-FNO, which uses an end-to-end approach to learn a shared deformation map and the latent operator, is less suited for the diverse shapes present in the FlowBench dataset. In contrast, our OTNO model, which solves the OT plan/map for each shape separately, exhibits superior performance on diverse geometries.

\begin{table}[h]
  \centering
  \caption{Prediction under \textbf{M1} Metric (global) on FlowBench Dataset (single A100)}
  \label{table:flowbench-m1}
  \centering
  \begin{tabular}{lcccc} 
    \toprule 
    \small{Group} & \small{Model}	&\small{Relative L2}   &\small{Time per Epoch (sec)}    & \small{GPU Memory (MB)}\\
    \midrule
    \multirow{3}{*}{G1}
        & FNO & 16.27\% & 43 & 4548  \\
        & GINO & 8.62\% & 371 & 57870 \\
        & OTNO & \textbf{3.06\%} & 578 & 26324 \\
    \midrule
    \multirow{3}{*}{G2} 
        & FNO & 56.67\% & 43 & 4548 \\
        & GINO & 43.16\% & 390 & 58970 \\
        & OTNO & \textbf{7.16\%} & 603 & 22868 \\
    \midrule
    \multirow{3}{*}{G3} 
        & FNO & 23.20\% & 44 & 4548 \\
        & GINO & 13.27\% & 383 & 73140 \\
        & OTNO & \textbf{4.02\%} & 606 & 26008 \\
    \bottomrule
  \end{tabular}
\end{table}

\begin{table}[h]
  \centering
  \caption{Prediction under \textbf{M2} Metric (boundary) on FlowBench Dataset (single A100)}
  \label{table:flowbench-m2}
  \centering
  \begin{tabular}{lcccc} 
    \toprule 
    \small{Group} & \small{Model}	&\small{Relative L2}   &\small{Time per Epoch (sec)}    & \small{GPU Memory (MB)}\\
    \midrule
    \multirow{3}{*}{G1} 
        & FNO & 5.65\% & 43 & 4536 \\ 
        & GINO & 5.83\% & 190 & 67632 \\ 
        & OTNO & \textbf{3.91\%} & 135 & 7300 \\
    \midrule
    \multirow{3}{*}{G2} 
        & FNO & 29.37\% & 43 & 4534 \\
        & GINO & 19.74\% & 177 & 73792 \\
        & OTNO & \textbf{14.36\%} & 124 & 7012 \\
    \midrule
    \multirow{3}{*}{G3} 
        & FNO & 10.47\% & 43 & 4538 \\
        & GINO & 10.69\% & 219 & 67838 \\
        & OTNO & \textbf{7.18\%} & 153 & 11406 \\
    \bottomrule
  \end{tabular}
\end{table}

\begin{figure}[h]
    \centering
    \begin{minipage}{1.0\textwidth}
        \centering
        \includegraphics[width=\linewidth]{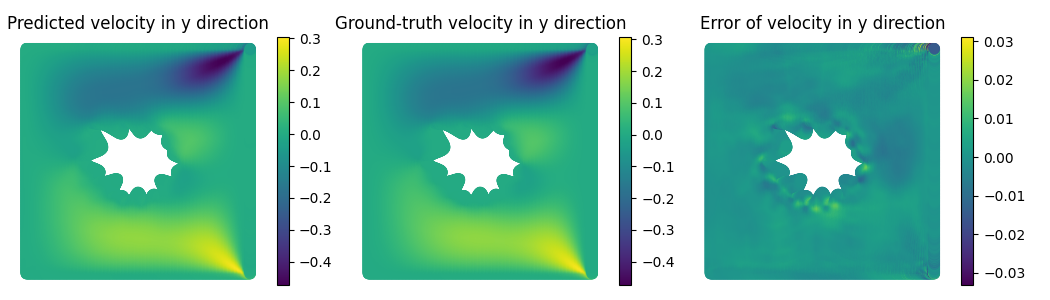}
        \subcaption{Results of velocity in y direction under M1 metric (global).}
    \end{minipage}\\
    \begin{minipage}{1.0\textwidth}
        \centering
        \includegraphics[width=\linewidth]{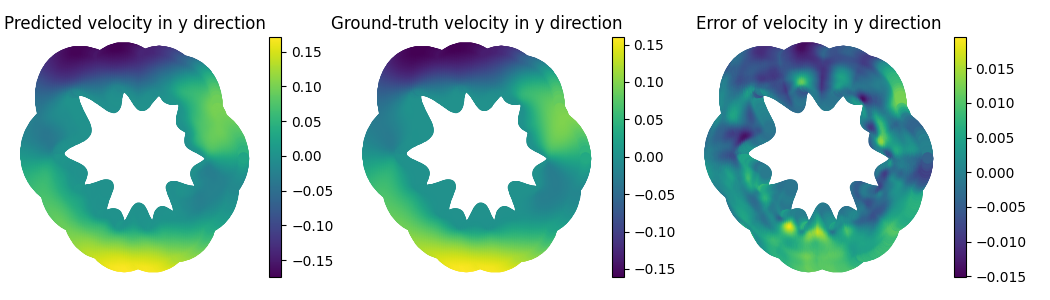}
        \subcaption{Results of velocity in y direction under M2 metric (boundary).}
    \end{minipage}\\
    \caption{Results Visualization for OTNO on FlowBench Dataset}
    \label{fig:flowbench_visualization}
\end{figure}
\section{Ablation Studies}

\subsection{OTNO - Plan (Sinkhorn)}
\subsubsection{Encoder \& Decoder Strategy}
\label{ot_strategy}
Using the Sinkhorn algorithm, we solve the Kantorovich optimal transport problem between the latent mesh $\Xi \in \mathbb{R}^{n_2}$ and the physical mesh $\mathcal{X} \in \mathbb{R}^{n_1}$, resulting in a large, dense coupling matrix $P \in \mathbb{R}^{n_2 \times n_1}$ that approximates the OT plan. The direct way to get transported mesh is as $\mathcal{X}' = P \cdot \mathcal{X}$, we refer to as the Matrix Strategy. However, storing these large, dense matrices for each physical mesh is too costly, and the approximate matrices obtained from the Sinkhorn Method introduce a degree of imprecision in the results. Therefore, this section discusses strategies to implement these approximate matrices more efficiently, aiming to conserve memory and reduce imprecision.

\begin{enumerate}
    \item \textbf{Max vs Mean}
    
As the matrix \(P\in \mathbb{R}^{n_2 \times n_1}\) represents a discrete probability measure on \(\Xi \times \mathcal{X}\), a natural approach to take use of the plan is to transport by the maximum probability, termed as "Max" strategy. 
Let \(\mathcal{X} = (x_1, \dots, x_{n_1}) \in \mathbb{R}^{n_1 \times d}\). Denote \( P = (p_{k,l})_{n_2 \times n_1}\). We calculate the indices of the max element in each row
\begin{equation}
i_k = \underset{j}{\arg \min} \{ p_{k,j} : j=1, \dots, n_1 \}, \text{for } k=1,\dots,n_2,
\end{equation}
and then use these indices to get a refined transported mesh: \(\mathcal{M} = (x_{i_1}, \dots, x_{i_{n_2}}) \in \mathbb{R}^{n_2 \times d}\).

Another approach, perhaps more intuitive, involves finding the element in mesh \(\mathcal{X}\) that is closest to the directly transported mesh \(\mathcal{X}' = (x_1',\dots,x_{n_2}') \in \mathbb{R}^{n_2 \times d}\). We name this approach as "Mean" strategy due to the weight product across rows in $\mathcal{X}'=P\cdot \mathcal{X}$. The specific process is described as follows: first compute the indices 
\begin{equation}
i_k = \underset{j}{\arg\min} \{ |x'_k - x_j|: j=1, \dots, n_1\}, \text{for } k=1,\dots,n_2.
\end{equation}
The refined transported mesh is \(\mathcal{M} = (x_{i_1}, \dots, x_{i_{n_2}}) \in \mathbb{R}^{n_2 \times d}\).

As shown in Table.\ref{tab:max_mean_comparison}, "Mean" strategy has a better performance.
\begin{table}[h]
\centering
\caption{Comparison of Max vs. Mean Strategy}
\label{tab:max_mean_comparison}
\begin{tabular}{lccc}
\toprule
\textbf{Dataset} & \textbf{Matrix} & \textbf{Max} & \textbf{Mean} \\
\midrule
ShapeNet (Relative L2) &7.21\% & 7.01\% & \textbf{6.70\%} \\
DrivAerNet (MSE e-5) &5.6 & 4.8 & \textbf{3.4} \\
\bottomrule
\end{tabular}
\end{table}

\item \textbf{Single vs Multiple}

Besides the implementation of OT, integrating OT with FNO poses another significant question. For FNO, the requirement of inputs is just need to be features on a latent grid, suggesting that the encoder's output can combine multiple transported distributions, same for the decoder. Thus, under the "Mean" strategy, we further investigate utilizing indices of the top \(k\) closest elements, termed as "Multi-Enc" and "Multi-Dec". We choose $k=8$ for "Multi-End" and $k=3$ for "Multi-Dec" setups. The comparative results are presented in Table \ref{tab:multi_simple_comparison}, indicating that "Multi-Enc" and "Multi-Dec" configurations underperform relative to "Single" strategy which utilize the index of the closest point instead of top $k$ closest points.

\begin{table}[h]
\centering
\caption{Comparison of Multi vs. Single Encoder/Decoder Strategy}
\label{tab:multi_simple_comparison}
\begin{tabular}{lccc}
\toprule
\textbf{Dataset} & \textbf{Multi Enc} & \textbf{Multi Dec} & \textbf{Single} \\
\midrule
ShapeNet (Relative L2) & 20.55\% & 72.30\% & \textbf{6.70\%} \\
DrivAerNet (MSE e-5) & 4.2 & 4.5 & \textbf{3.4} \\
\bottomrule
\end{tabular}
\end{table}
\end{enumerate}

\subsubsection{Normal Features}
\label{normal features}
Our model, which incorporates latent FNO, learns the operator mapping from \(T\) to the latent solution function \(v\), where \(T\) represents the transport map or the marginal map of the transport plan, as described in equations \eqref{latent operator - map} and \eqref{latent operator - plan}. The basic representation of the map \(T\) in our experiments can be defined as \(\mathcal{T} = (\Xi, T(\Xi))\). The previous section discussed how to refine the transported mesh \(T(\Xi)\), and in this section, we further explore the addition of normal features to enhance the representation \(\mathcal{T}\), leveraging the normal's ability to describe a surface.

We propose three different approaches to integrate normals and compare them with the case where no normal features are added. The three methods are as follows: "Car": Only add car surface normals. "Concatenate": Add the concatenation of latent surface normals and car surface normals as normal features. "Cross Product": Add the cross product of latent surface normals and car surface normals as normal features to capture the deformation information of the transport.

As shown in Table~\ref{tab:normal_features}, the "Cross Product" method performs the best.
\begin{table}[h]
\centering
\caption{Comparison of Different Normal Features}
\label{tab:normal_features}
\begin{tabular}{lcccc}
\toprule
\textbf{Dataset} & \textbf{Non} & \textbf{Car} & \textbf{Concatenate} & \textbf{Cross Product} \\
\midrule
ShapeNet (Rel L2) & 7.19\% & 6.82\% & 6.83\% & \textbf{6.70\%} \\
DrivAerNet (MSE e-5) & 3.4  & 3.9 & 3.8 & \textbf{3.4}  \\
\bottomrule
\end{tabular}
\end{table}

\subsubsection{Shape of mesh in latent space}
\label{latent shape}
We investigated the effects of different shapes for the latent mesh, i.e., the target shapes for optimal transport. The results presented in Table~\ref{tab:mesh_shape} indicate that the torus provides the best performance due to its alignment with the periodic Fourier function. Although the capped hemisphere and the spherical surface share the same topological structure as the car surface, their performance is suboptimal. Additionally, we experimented with the double sphere method \citep{mildenberger2023double}, which unfortunately yielded worse accuracy compared to the sphere and doubled the time costs.

\begin{table}[h]
\centering
\caption{Comparison of Different Mesh Shapes in Latent Space}
\label{tab:mesh_shape}
\begin{tabular}{lccccc}
\toprule
\textbf{Dataset} & \textbf{Hemisphere} & \textbf{Plane} & \textbf{Double Sphere} & \textbf{Sphere} & \textbf{Torus} \\
\midrule
ShapeNet (Rel L2) & 8.9\% & 8.67\% & 7.41\% & 7.09\% & \textbf{6.70\%} \\
DrivAerNet (MSE e-5) & 4.7  & 4.4 & 4.6 & 4.1 & \textbf{3.4}  \\
\bottomrule
\end{tabular}
\end{table}

\subsubsection{Expand Factor}
\label{latent size}
\begin{figure}[h]
    \centering
    \begin{minipage}{0.48\textwidth}
        \centering
        \includegraphics[width=\linewidth]{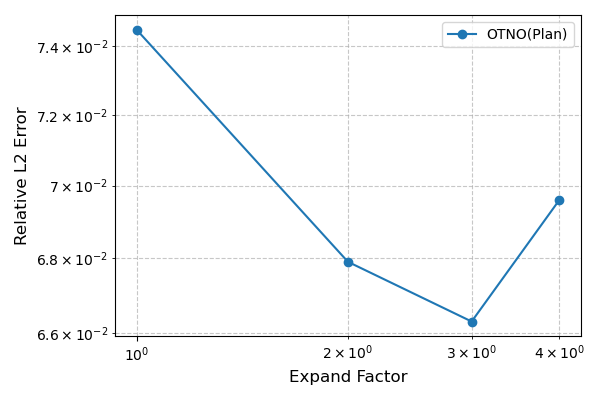}
        \subcaption{Tests of different expand factor for OTNO(Plan) on ShapeNet dataset
        }
        \label{fig:shapenet-latentsize}
    \end{minipage}\hfill
    \begin{minipage}{0.48\textwidth}
        \centering
        \includegraphics[width=\linewidth]{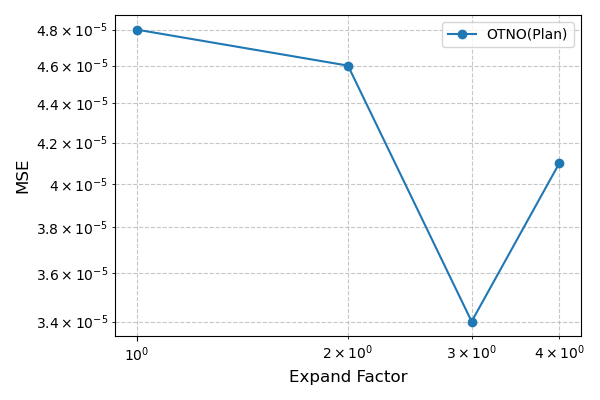}
        \subcaption{Tests of different expand factor for OTNO(Plan) on DrivAerNet dataset
        }
        \label{fig:drivaernet-latentsize}
    \end{minipage}
    \caption{Ablation Studies of expand factor for OTNO(Plan)
    }
    \label{fig:expand_factor}
\end{figure}

For OTNO(Plan), we found that expanding the size of the latent mesh within a certain range leads to better results. The ablation experiments for different expansion factors $\alpha=1,2,3,4$are presented. As shown in Fig.~\ref{fig:expand_factor}, on both the DrivAerNet and ShapeNet datasets, $\alpha =3$ achieves the best accuracy.

\subsection{OTNO - Map (PPMM)}
\subsubsection{Number of Iterations}
The theoretical time complexity of the PPMM is \( O(Kn\log(n)) \), where \( K \) is the number of iterations and \( n \) is the number of points. While the original study claims that empirically \( K = O(d) \) works reasonably well, our experience with 3D car datasets tells a different story. In this section, we conduct ablation studies to explore the relationship between the optimal number of iterations and the number of points. We employ voxel downsampling to generate subsets of the car surface mesh, varying in the number of points. The results for different voxel sizes (\( r \)) and iteration numbers (\( K \)) are presented in Table~\ref{tab:ppmm-itr}. From these results, we observe that \( K \propto r^{-1} \). Given that the car surface mesh is a 2D manifold embedded in 3D space, the number of points \( n \) should be inversely proportional to the square of the voxel size \( r \), i.e., \( n \propto r^{-2} \). Consequently, \( K \propto n^{1/2} \), and the experimental time complexity of PPMM in our model on the car dataset is accordingly \( O(n^{3/2} \log(n)) \).

\begin{table}[h]
\centering
\caption{MSE (e-5) Results on the DrivAerNet Dataset for Different Voxel Sizes and Different Iteration Numbers of PPMM}
\label{tab:ppmm-itr}
\begin{tabular}{lcccc}
\toprule
\textbf{Voxel Size / Itr} & 500 & 1000 & 2000 & 4000 \\
\midrule
0.2 & \textbf{6.2} & 6.6 & 6.9 & 8.0 \\
0.1 & 5.3 & \textbf{4.6} & 5.5 & 5.4 \\
0.05 & 4.8 & 4.2 & \textbf{3.9} & 4.5 \\
\bottomrule
\end{tabular}
\end{table}

\subsubsection{Sampling Mesh Size}
\label{ablation: physical sampling}
It is a challenging problem to efficiently and effectively solve large-scale OT problems. We investigate the Sinkhorn method for large-scale OT plans and the PPMM algorithm for large-scale OT maps, and accordingly build two models, OTNO(Plan) and OTNO(Map). In this section, we conduct experiments on the large-scale DrivAerNet dataset to explore our models' ability to handle large sampling meshes. We use voxel downsampling to reduce the mesh with 200k points from the DrivAerNet dataset into a normalized sampling mesh of varying sizes.

\begin{figure}[h]
    \centering
    \begin{minipage}{0.48\textwidth}
        \centering
        \includegraphics[width=\linewidth]{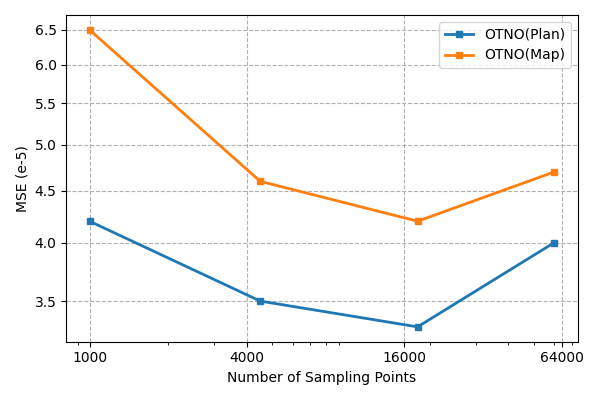}
        \subcaption{Test errors associate with the number of sampling points. We set the voxel size $r=0.2,0.1,0.05,0.025$, corresponding to about 1k, 4.5k, 18k, 60k points.
        }
        \label{fig:samplingsize-errors}
    \end{minipage}\hfill
    \begin{minipage}{0.48\textwidth}
        \centering
        \includegraphics[width=\linewidth]{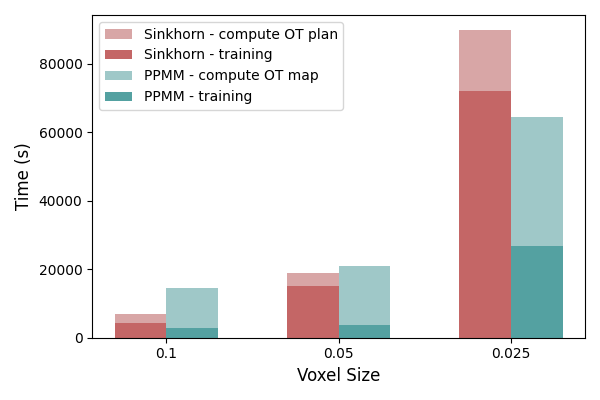}
        \subcaption{Time complexity associate with the number of sampling points. We set the voxel size $r=0.1,0.05,0.025$, corresponding to about 4.5k, 18k, 60k points.}
        \label{fig:samplingsize-times}
    \end{minipage}
    \caption{Ablation Studies of Sampling Mesh Size for OTNO(Map) and OTNO(Plan) on DrivAerNet Dataset
    }
    \label{fig:ablation-map}
\end{figure}

As shown in Fig.~\ref{fig:samplingsize-errors}, both OTNO(Plan) and OTNO(Map) achieve their lowest accuracy at a voxel size of 0.05, corresponding to approximately 18k points. However, further increasing the sampling mesh size does not lead to improved accuracy. This suggests that our model has a range of limitations when dealing with large-scale problems, primarily due to the difficulty of solving large-scale OT problems with high precision. Regarding the comparison between OTNO(Plan) and OTNO(Map), we find that OTNO(Plan) consistently outperforms OTNO(Map) in terms of accuracy, regardless of the sampling mesh size.

From the experimental results, we observe that the training time of OTNO(Plan) is much larger than OTNO(Map) as shown in Fig.~\ref{fig:samplingsize-times}. This is because we expand the latent space by a factor of 3 for OTNO(Plan), and the FNO on the latent space has linear complexity, attributed to the linear FFT. Note that the time complexity of Sinkhorn and PPMM are \(O(n^2)\) and \(O(n^{3/2} \log(n))\), respectively, both of which are larger than the FNO training complexity of \(O(n)\). Therefore, the overall time complexity of OTNO(Plan) and OTNO(Map) relative to the number of points \(n\) should be \(O(n^2)\) and \(O(n^{3/2} \log(n))\), respectively. These results align with the plots shown in Fig.~\ref{fig:samplingsize-times} that the time cost of OTNO(Plan) increases faster than that of OTNO(Map) as the sampling mesh size grows.

\section{Discussion}
\subsection{Conformal Mapping}
Conformal mapping \citep{gu2004genus, choi2015flash} represents a common approach for mapping irregular meshes to canonical manifolds. These mappings preserve local angles, making them particularly suitable for Laplacian-type equations. The smoothness and harmonic properties of conformal mappings work exceptionally well with numerical solvers such as finite element methods, as they have less local distortion on each cell within a mesh.
However, despite these advantages, conformal mapping is suboptimal for evaluating the integral operators widely employed in neural operators. For this reason, our work adopts optimal transport as the preferred geometric transformation.  While optimal transport may not be as smooth locally, they preserve global measure, which is essential for integral operators.

To test this, we conducted a preliminary experiment on the ShapeNet-Car dataset. We used conformal mapping to embed the car surface into a latent spherical mesh and then applied a neural operator in this latent space. We refer to this model as the Conformal Mapping Neural Operator (CMNO). As shown in Table~\ref{tab:conformal_mapping}, OTNO significantly outperforms CMNO. 

\begin{table}[h]
  \caption{Predict pressure field on ShapeNet Car Dataset (single P100)}
  \label{tab:conformal_mapping}
  \centering
  \begin{tabular}{lccc} 
    \toprule
    Model    & Relative L2   & Total Time (hr) \\
    \midrule
    CMNO & 12.79\% & 1.13 & \\
    OTNO(Plan)   & 6.70\%   & 1.52 \\   
    \bottomrule
  \end{tabular}
\end{table}

\subsection{diffeomorphic transformations}
Recently, several studies have focused on integrating diffeomorphic transformations within PDE solution operators to extract geometric information and ultimately solve PDEs. Examples include DIMON \citep{yin2024dimon}, which employs LDDMM to learn diffeomorphic deformations, and DNO \citep{zhao2025DNO}, which utilizes harmonic mapping for the same purpose.

A recent work, Diffeomorphic Latent Neural Operators \citep{ahmad2024diffeomorphiclatentneuraloperators}, compared conformal mapping and LDDMM, which are both diffeomorphic mappings, with discrete optimal transport via the Hungarian algorithm, which is a non-diffeomorphic mapping.  Their findings suggest that, for invertible spatial transformations, diffeomorphic approaches generally offer better performance.

However, the Hungarian algorithm is designed primarily for discrete assignment problems, which is not suited for spatial transformations between meshes. In contrast, our method employs continuous OT algorithms, which are better equipped for distribution transformations that encode/decode geometric information.

We observe several advantages of our method employing continuous OT algorithms compared to diffeomorphic methods. First, we found that smoothness is an unnecessarily strict requirement—piecewise continuity is often sufficient for effective geometric embeddings. For example, using a sphere surface as a latent shape results in a globally continuous transported mesh, whereas for a torus, the transported mesh is only piecewise continuous. Yet, as shown in Table~\ref{tab:mesh_shape}, the torus-based representation outperforms the sphere-based one. Secondly invertibility is not always beneficial. Invertibility requires the size of latent mesh should be the same as the size of phyical mesh. However, As illustrated in Fig.~\ref{fig:shapenet-latentsize}, expanding the size of latent mesh within a certain range leads to improved performance.

Moreover, the topology of the mapping warrants attention. While diffeomorphisms preserve topology, our OT framework does not require topological consistency. As demonstrated in Table~\ref{tab:mesh_shape}, the torus representation achieves the best performance despite a change in topology. We also tested this on a 2D elasticity example to assess its generalization capability. The results, shown in Fig.~\ref{topology}, confirm that enforcing topological consistency does not yield improved outcomes.

\begin{table}[ht]
  \caption{Demonstrate topology independence of OTNO on elasticity}
  \label{topology}
  \centering
  \begin{tabular}{llll} 
    \toprule
    Latent grid   & Topology    & Relative Error \\
    \midrule
    Ring  & \usym{2714}   & 3.8\% \\
    Square   & \usym{2715}  & 2.7\% \\
    
    \bottomrule
  \end{tabular}
\end{table}

\section{Conclusion}
In this work, we propose an OT framework for geometry embedding, which maps density functions from physical geometric domains to latent uniform density functions on regular latent geometric domains. We developed the OTNO model by integrating neural operator with optimal transport to solve PDEs on complex geometries. Specifically, there are two implementations: OTNO(Plan) and OTNO(Map), using Kantorovich and Monge formulations, respectively.

Our model achieves particularly good performance in automotive and aerospace applications, where the inputs are 2D surface designs, and the outputs are surface pressure and velocity. These applications provide the opportunity to use optimal transport to embed the physical surface mesh in 3D space into a 2D parameterized latent mesh, allowing computations in a lower-dimensional space. The effectiveness and efficiency of our model in these scenarios are confirmed by our experiments on the ShapeNet-Car and DrivAerNet-Car datasets.

Moreover, by leveraging the advantage of instance-dependent OT Map/Plan, our model handles diverse geometries effectively. As demonstrated on the FlowBench dataset, which includes a wider variety of shapes, OTNO significantly outperforms other models in terms of accuracy.

While our proposed methods demonstrate promise, they present certain limitations that pave the way for future research. A primary challenge lies in the trade-off between computational complexity and accuracy. The Sinkhorn algorithm, underpinning our OTNO(Plan) model, exhibits a complexity of $O(n^2)$, posing scalability challenges for large-scale point clouds. Our alternative, the OTNO(Map) model leveraging the PPMM method, achieves a reduced experimental complexity of $O(n^{3/2}\log(n))$ in a lower-dimensional Euclidean space, yet at the cost of diminished accuracy compared to OTNO(Plan). Consequently, a key direction for future work is the development of novel, sub-quadratic algorithms for optimal transport that can achieve higher accuracy, thereby bridging this performance gap.

A second promising avenue involves a deeper investigation into the optimal selection of the latent sub-manifold. Our experimental results (cf. Table \ref{tab:mesh_shape}) indicate that the optimal latent manifold is non-canonical, diverging from the intuitively favored spherical topology typically chosen for its topological preservation properties. Future efforts will focus on systematically exploring criteria and methods for determining the most suitable latent manifold structure to enhance model performance.

\acks{{Anima Anandkumar is supported by the Bren named chair professorship, Schmidt AI2050 senior fellowship, and ONR (MURI grant N00014-18-1-2624). }}

\vskip 0.2in
\bibliography{reference}

\begin{thebibliography}{51}
\providecommand{\natexlab}[1]{#1}
\providecommand{\url}[1]{\texttt{#1}}
\expandafter\ifx\csname urlstyle\endcsname\relax
  \providecommand{\doi}[1]{doi: #1}\else
  \providecommand{\doi}{doi: \begingroup \urlstyle{rm}\Url}\fi

\bibitem[Ahmad et~al.(2024)Ahmad, Chen, Yin, Kumar, Charon, Trayanova, and Maggioni]{ahmad2024diffeomorphiclatentneuraloperators}
Zan Ahmad, Shiyi Chen, Minglang Yin, Avisha Kumar, Nicolas Charon, Natalia Trayanova, and Mauro Maggioni.
\newblock Diffeomorphic latent neural operators for data-efficient learning of solutions to partial differential equations, 2024.
\newblock URL \url{https://arxiv.org/abs/2411.18014}.

\bibitem[Alkin et~al.(2024)Alkin, F{\"u}rst, Schmid, Gruber, Holzleitner, and Brandstetter]{alkin2024universal}
Benedikt Alkin, Andreas F{\"u}rst, Simon Schmid, Lukas Gruber, Markus Holzleitner, and Johannes Brandstetter.
\newblock Universal physics transformers.
\newblock \emph{arXiv preprint arXiv:2402.12365}, 2024.

\bibitem[Bhatnagar et~al.(2019)Bhatnagar, Afshar, Pan, Duraisamy, and Kaushik]{bhatnagar2019prediction}
Saakaar Bhatnagar, Yaser Afshar, Shaowu Pan, Karthik Duraisamy, and Shailendra Kaushik.
\newblock Prediction of aerodynamic flow fields using convolutional neural networks.
\newblock \emph{Computational Mechanics}, 64:\penalty0 525--545, 2019.

\bibitem[Bonev et~al.(2023)Bonev, Kurth, Hundt, Pathak, Baust, Kashinath, and Anandkumar]{bonev2023spherical}
Boris Bonev, Thorsten Kurth, Christian Hundt, Jaideep Pathak, Maximilian Baust, Karthik Kashinath, and Anima Anandkumar.
\newblock Spherical fourier neural operators: Learning stable dynamics on the sphere.
\newblock In \emph{International conference on machine learning}, pages 2806--2823. PMLR, 2023.

\bibitem[Brenier(1991)]{brenier1991polar}
Yann Brenier.
\newblock Polar factorization and monotone rearrangement of vector-valued functions.
\newblock \emph{Communications on pure and applied mathematics}, 44\penalty0 (4):\penalty0 375--417, 1991.

\bibitem[Budd et~al.(2015)Budd, Russell, and Walsh]{budd2015geometry}
CJ~Budd, Robert~D Russell, and E~Walsh.
\newblock The geometry of r-adaptive meshes generated using optimal transport methods.
\newblock \emph{Journal of Computational Physics}, 282:\penalty0 113--137, 2015.

\bibitem[Chang et~al.(2015)Chang, Funkhouser, Guibas, Hanrahan, Huang, Li, Savarese, Savva, Song, Su, et~al.]{chang2015shapenet}
Angel~X Chang, Thomas Funkhouser, Leonidas Guibas, Pat Hanrahan, Qixing Huang, Zimo Li, Silvio Savarese, Manolis Savva, Shuran Song, Hao Su, et~al.
\newblock Shapenet: An information-rich 3d model repository.
\newblock In \emph{arXiv:1512.03012 [cs.GR]}, 2015.

\bibitem[Chen et~al.(2023)Chen, Wu, Grinspun, Zheng, and Chen]{chen2023implicit}
Honglin Chen, Rundi Wu, Eitan Grinspun, Changxi Zheng, and Peter~Yichen Chen.
\newblock Implicit neural spatial representations for time-dependent pdes.
\newblock In \emph{International Conference on Machine Learning}, pages 5162--5177. PMLR, 2023.

\bibitem[Chen et~al.(2022)Chen, Xiang, Cho, Chang, Pershing, Maia, Chiaramonte, Carlberg, and Grinspun]{chen2022crom}
Peter~Yichen Chen, Jinxu Xiang, Dong~Heon Cho, Yue Chang, GA~Pershing, Henrique~Teles Maia, Maurizio~M Chiaramonte, Kevin Carlberg, and Eitan Grinspun.
\newblock Crom: Continuous reduced-order modeling of pdes using implicit neural representations.
\newblock \emph{arXiv preprint arXiv:2206.02607}, 2022.

\bibitem[Choi et~al.(2015)Choi, Lam, and Lui]{choi2015flash}
Pui~Tung Choi, Ka~Chun Lam, and Lok~Ming Lui.
\newblock Flash: Fast landmark aligned spherical harmonic parameterization for genus-0 closed brain surfaces.
\newblock \emph{SIAM Journal on Imaging Sciences}, 8\penalty0 (1):\penalty0 67--94, 2015.

\bibitem[Choy et~al.(2025)Choy, Kamenev, Kossaifi, Rietmann, Kautz, and Azizzadenesheli]{choy2025factorized}
Chris Choy, Alexey Kamenev, Jean Kossaifi, Max Rietmann, Jan Kautz, and Kamyar Azizzadenesheli.
\newblock Factorized implicit global convolution for automotive computational fluid dynamics prediction.
\newblock \emph{arXiv preprint arXiv:2502.04317}, 2025.

\bibitem[Cloud(2025)]{shift_suv_2025}
Luminary Cloud.
\newblock Shift-suv sample: High-fidelity computational fluid dynamics dataset for suv external aerodynamics, 2025.
\newblock URL \url{https://huggingface.co/datasets/LuminaryCloud/shift-suv-samples}.

\bibitem[Cooley and Tukey(1965)]{FFT}
James~W Cooley and John~W Tukey.
\newblock An algorithm for the machine calculation of complex fourier series.
\newblock \emph{Mathematics of computation}, 19\penalty0 (90):\penalty0 297--301, 1965.

\bibitem[Cuturi(2013)]{cuturi2013sinkhorn}
Marco Cuturi.
\newblock Sinkhorn distances: Lightspeed computation of optimal transport.
\newblock \emph{Advances in neural information processing systems}, 26, 2013.

\bibitem[Elrefaie et~al.(2024{\natexlab{a}})Elrefaie, Dai, and Ahmed]{elrefaie2024drivaernetparametriccardataset}
Mohamed Elrefaie, Angela Dai, and Faez Ahmed.
\newblock Drivaernet: A parametric car dataset for data-driven aerodynamic design and graph-based drag prediction, 2024{\natexlab{a}}.
\newblock URL \url{https://arxiv.org/abs/2403.08055}.

\bibitem[Elrefaie et~al.(2024{\natexlab{b}})Elrefaie, Morar, Dai, and Ahmed]{elrefaie2024drivaernet++}
Mohamed Elrefaie, Florin Morar, Angela Dai, and Faez Ahmed.
\newblock Drivaernet++: A large-scale multimodal car dataset with computational fluid dynamics simulations and deep learning benchmarks.
\newblock \emph{arXiv preprint arXiv:2406.09624}, 2024{\natexlab{b}}.

\bibitem[Flamary et~al.(2021)Flamary, Courty, Gramfort, Alaya, Boisbunon, Chambon, Chapel, Corenflos, Fatras, Fournier, Gautheron, Gayraud, Janati, Rakotomamonjy, Redko, Rolet, Schutz, Seguy, Sutherland, Tavenard, Tong, and Vayer]{flamary2021pot}
R{\'e}mi Flamary, Nicolas Courty, Alexandre Gramfort, Mokhtar~Z. Alaya, Aur{\'e}lie Boisbunon, Stanislas Chambon, Laetitia Chapel, Adrien Corenflos, Kilian Fatras, Nemo Fournier, L{\'e}o Gautheron, Nathalie~T.H. Gayraud, Hicham Janati, Alain Rakotomamonjy, Ievgen Redko, Antoine Rolet, Antony Schutz, Vivien Seguy, Danica~J. Sutherland, Romain Tavenard, Alexander Tong, and Titouan Vayer.
\newblock Pot: Python optimal transport.
\newblock \emph{Journal of Machine Learning Research}, 22\penalty0 (78):\penalty0 1--8, 2021.
\newblock URL \url{http://jmlr.org/papers/v22/20-451.html}.

\bibitem[Gu et~al.(2004)Gu, Wang, Chan, Thompson, and Yau]{gu2004genus}
Xianfeng Gu, Yalin Wang, Tony~F Chan, Paul~M Thompson, and Shing-Tung Yau.
\newblock Genus zero surface conformal mapping and its application to brain surface mapping.
\newblock \emph{IEEE transactions on medical imaging}, 23\penalty0 (8):\penalty0 949--958, 2004.

\bibitem[Hao et~al.(2023)Hao, Wang, Su, Ying, Dong, Liu, Cheng, Song, and Zhu]{hao2023gnot}
Zhongkai Hao, Zhengyi Wang, Hang Su, Chengyang Ying, Yinpeng Dong, Songming Liu, Ze~Cheng, Jian Song, and Jun Zhu.
\newblock Gnot: A general neural operator transformer for operator learning.
\newblock In \emph{International Conference on Machine Learning}, pages 12556--12569. PMLR, 2023.

\bibitem[Hennigh et~al.(2021)Hennigh, Narasimhan, Nabian, Subramaniam, Tangsali, Fang, Rietmann, Byeon, and Choudhry]{hennigh2021nvidia}
Oliver Hennigh, Susheela Narasimhan, Mohammad~Amin Nabian, Akshay Subramaniam, Kaustubh Tangsali, Zhiwei Fang, Max Rietmann, Wonmin Byeon, and Sanjay Choudhry.
\newblock Nvidia simnet™: An ai-accelerated multi-physics simulation framework.
\newblock In \emph{International conference on computational science}, pages 447--461. Springer, 2021.

\bibitem[Hubbert(1956)]{hubbert1956darcy}
M~King Hubbert.
\newblock Darcy's law and the field equations of the flow of underground fluids.
\newblock \emph{Transactions of the AIME}, 207\penalty0 (01):\penalty0 222--239, 1956.

\bibitem[Ihlenburg and Babu{\v{s}}ka(1995)]{Helmholtz}
Frank Ihlenburg and Ivo Babu{\v{s}}ka.
\newblock Finite element solution of the helmholtz equation with high wave number part i: The h-version of the fem.
\newblock \emph{Computers \& Mathematics with Applications}, 30\penalty0 (9):\penalty0 9--37, 1995.

\bibitem[Kantorovich(2006)]{kantorovich2006problem}
LK~Kantorovich.
\newblock On a problem of monge.
\newblock \emph{Journal of Mathematical Sciences}, 133\penalty0 (4), 2006.

\bibitem[Kovachki et~al.(2023)Kovachki, Li, Liu, Azizzadenesheli, Bhattacharya, Stuart, and Anandkumar]{kovachki2023neural}
Nikola Kovachki, Zongyi Li, Burigede Liu, Kamyar Azizzadenesheli, Kaushik Bhattacharya, Andrew Stuart, and Anima Anandkumar.
\newblock Neural operator: Learning maps between function spaces with applications to pdes.
\newblock \emph{Journal of Machine Learning Research}, 24\penalty0 (89):\penalty0 1--97, 2023.

\bibitem[Li et~al.(2019)Li, Muller, Thabet, and Ghanem]{li2019deepgcns}
Guohao Li, Matthias Muller, Ali Thabet, and Bernard Ghanem.
\newblock Deepgcns: Can gcns go as deep as cnns?
\newblock In \emph{Proceedings of the IEEE/CVF international conference on computer vision}, pages 9267--9276, 2019.

\bibitem[Li et~al.(2020{\natexlab{a}})Li, Kovachki, Azizzadenesheli, Liu, Bhattacharya, Stuart, and Anandkumar]{FNO}
Zongyi Li, Nikola Kovachki, Kamyar Azizzadenesheli, Burigede Liu, Kaushik Bhattacharya, Andrew Stuart, and Anima Anandkumar.
\newblock Fourier neural operator for parametric partial differential equations.
\newblock \emph{arXiv preprint arXiv:2010.08895}, 2020{\natexlab{a}}.

\bibitem[Li et~al.(2020{\natexlab{b}})Li, Kovachki, Azizzadenesheli, Liu, Bhattacharya, Stuart, and Anandkumar]{li2020neural}
Zongyi Li, Nikola Kovachki, Kamyar Azizzadenesheli, Burigede Liu, Kaushik Bhattacharya, Andrew Stuart, and Anima Anandkumar.
\newblock Neural operator: Graph kernel network for partial differential equations.
\newblock \emph{arXiv preprint arXiv:2003.03485}, 2020{\natexlab{b}}.

\bibitem[Li et~al.(2021)Li, Kovachki, Azizzadenesheli, Liu, Bhattacharya, Stuart, and Anandkumar]{GNO}
Zongyi Li, Nikola Kovachki, Kamyar Azizzadenesheli, Burigede Liu, Kaushik Bhattacharya, Andrew Stuart, and Anima Anandkumar.
\newblock Neural operator: Graph kernel network for partial differential equations.
\newblock \emph{ICLR}, 2021.

\bibitem[Li et~al.(2023{\natexlab{a}})Li, Huang, Liu, and Anandkumar]{Geo-FNO}
Zongyi Li, Daniel~Zhengyu Huang, Burigede Liu, and Anima Anandkumar.
\newblock Fourier neural operator with learned deformations for pdes on general geometries.
\newblock \emph{Journal of Machine Learning Research}, 24\penalty0 (388):\penalty0 1--26, 2023{\natexlab{a}}.

\bibitem[Li et~al.(2023{\natexlab{b}})Li, Kovachki, Choy, Li, Kossaifi, Otta, Nabian, Stadler, Hundt, Azizzadenesheli, and Anandkumar]{GINO}
Zongyi Li, Nikola~Borislavov Kovachki, Chris Choy, Boyi Li, Jean Kossaifi, Shourya~Prakash Otta, Mohammad~Amin Nabian, Maximilian Stadler, Christian Hundt, Kamyar Azizzadenesheli, and Anima Anandkumar.
\newblock Geometry-informed neural operator for large-scale 3d {PDE}s.
\newblock \emph{NIPS}, 2023{\natexlab{b}}.

\bibitem[Lu et~al.(2021)Lu, Jin, Pang, Zhang, and Karniadakis]{lu2021learning}
Lu~Lu, Pengzhan Jin, Guofei Pang, Zhongqiang Zhang, and George~Em Karniadakis.
\newblock Learning nonlinear operators via deeponet based on the universal approximation theorem of operators.
\newblock \emph{Nature machine intelligence}, 3\penalty0 (3):\penalty0 218--229, 2021.

\bibitem[Meng et~al.(2019)Meng, Ke, Zhang, Zhang, Zhong, and Ma]{meng2019large}
Cheng Meng, Yuan Ke, Jingyi Zhang, Mengrui Zhang, Wenxuan Zhong, and Ping Ma.
\newblock Large-scale optimal transport map estimation using projection pursuit.
\newblock \emph{Advances in Neural Information Processing Systems}, 32, 2019.

\bibitem[Mildenberger and Quellmalz(2023)]{mildenberger2023double}
Sophie Mildenberger and Michael Quellmalz.
\newblock A double fourier sphere method for d-dimensional manifolds.
\newblock \emph{Sampling Theory, Signal Processing, and Data Analysis}, 21\penalty0 (2):\penalty0 23, 2023.

\bibitem[Monge(1781)]{monge1781memoire}
Gaspard Monge.
\newblock M{\'e}moire sur la th{\'e}orie des d{\'e}blais et des remblais.
\newblock \emph{Mem. Math. Phys. Acad. Royale Sci.}, pages 666--704, 1781.

\bibitem[Peyr{\'e} et~al.(2019)Peyr{\'e}, Cuturi, et~al.]{peyre2019computational}
Gabriel Peyr{\'e}, Marco Cuturi, et~al.
\newblock Computational optimal transport: With applications to data science.
\newblock \emph{Foundations and Trends{\textregistered} in Machine Learning}, 11\penalty0 (5-6):\penalty0 355--607, 2019.

\bibitem[Pfaff et~al.(2021)Pfaff, Fortunato, Sanchez-Gonzalez, and Battaglia]{pfaff2020learning}
Tobias Pfaff, Meire Fortunato, Alvaro Sanchez-Gonzalez, and Peter~W Battaglia.
\newblock Learning mesh-based simulation with graph networks.
\newblock \emph{ICLR}, 2021.

\bibitem[Qi et~al.(2017)Qi, Yi, Su, and Guibas]{qi2017pointnet++}
Charles~Ruizhongtai Qi, Li~Yi, Hao Su, and Leonidas~J Guibas.
\newblock Pointnet++: Deep hierarchical feature learning on point sets in a metric space.
\newblock \emph{Advances in neural information processing systems}, 30, 2017.

\bibitem[Qian et~al.(2021)Qian, Hammoud, Li, Thabet, and Ghanem]{qian2021assanet}
Guocheng Qian, Hasan Hammoud, Guohao Li, Ali Thabet, and Bernard Ghanem.
\newblock Assanet: An anisotropic separable set abstraction for efficient point cloud representation learning.
\newblock \emph{Advances in Neural Information Processing Systems}, 34:\penalty0 28119--28130, 2021.

\bibitem[Qian et~al.(2022)Qian, Li, Peng, Mai, Hammoud, Elhoseiny, and Ghanem]{qian2022pointnext}
Guocheng Qian, Yuchen Li, Houwen Peng, Jinjie Mai, Hasan Hammoud, Mohamed Elhoseiny, and Bernard Ghanem.
\newblock Pointnext: Revisiting pointnet++ with improved training and scaling strategies.
\newblock \emph{Advances in neural information processing systems}, 35:\penalty0 23192--23204, 2022.

\bibitem[Santambrogio(2015)]{santambrogio2015optimal}
Filippo Santambrogio.
\newblock \emph{Optimal transport for applied mathematicians}, volume~87.
\newblock Springer, 2015.

\bibitem[Serrano et~al.(2023)Serrano, Le~Boudec, Kassa{\"\i}~Koupa{\"\i}, Wang, Yin, Vittaut, and Gallinari]{serrano2023operator}
Louis Serrano, Lise Le~Boudec, Armand Kassa{\"\i}~Koupa{\"\i}, Thomas~X Wang, Yuan Yin, Jean-No{\"e}l Vittaut, and Patrick Gallinari.
\newblock Operator learning with neural fields: Tackling pdes on general geometries.
\newblock \emph{Advances in Neural Information Processing Systems}, 36:\penalty0 70581--70611, 2023.

\bibitem[Stein(1970)]{stein1970singular}
Elias~M. Stein.
\newblock \emph{Singular Integrals and Differentiability Properties of Functions (PMS-30)}.
\newblock Princeton University Press, 1970.

\bibitem[Tali et~al.(2024)Tali, Rabeh, Yang, Shadkhah, Karki, Upadhyaya, Dhakshinamoorthy, Saadati, Sarkar, Krishnamurthy, et~al.]{tali2024flowBench}
Ronak Tali, Ali Rabeh, Cheng-Hau Yang, Mehdi Shadkhah, Samundra Karki, Abhisek Upadhyaya, Suriya Dhakshinamoorthy, Marjan Saadati, Soumik Sarkar, Adarsh Krishnamurthy, et~al.
\newblock Flowbench: A large scale benchmark for flow simulation over complex geometries.
\newblock \emph{arXiv preprint arXiv:2409.18032}, 2024.

\bibitem[Thuerey et~al.(2020)Thuerey, Wei{\ss}enow, Prantl, and Hu]{thuerey2020deep}
Nils Thuerey, Konstantin Wei{\ss}enow, Lukas Prantl, and Xiangyu Hu.
\newblock Deep learning methods for reynolds-averaged navier--stokes simulations of airfoil flows.
\newblock \emph{AIAA Journal}, 58\penalty0 (1):\penalty0 25--36, 2020.

\bibitem[Timmer(2009)]{timmer2009overview}
W~Timmer.
\newblock An overview of naca 6-digit airfoil series characteristics with reference to airfoils for large wind turbine blades.
\newblock In \emph{47th AIAA aerospace sciences meeting including the new horizons forum and aerospace exposition}, page 268, 2009.

\bibitem[Umetani and Bickel(2018)]{umetani2018learning}
Nobuyuki Umetani and Bernd Bickel.
\newblock Learning three-dimensional flow for interactive aerodynamic design.
\newblock \emph{ACM Transactions on Graphics (TOG)}, 37\penalty0 (4):\penalty0 1--10, 2018.

\bibitem[Wu et~al.(2024)Wu, Luo, Wang, Wang, and Long]{wu2024transolver}
Haixu Wu, Huakun Luo, Haowen Wang, Jianmin Wang, and Mingsheng Long.
\newblock Transolver: A fast transformer solver for pdes on general geometries.
\newblock \emph{arXiv preprint arXiv:2402.02366}, 2024.

\bibitem[Yin et~al.(2024)Yin, Charon, Brody, Lu, Trayanova, and Maggioni]{yin2024dimon}
Minglang Yin, Nicolas Charon, Ryan Brody, Lu~Lu, Natalia Trayanova, and Mauro Maggioni.
\newblock Dimon: Learning solution operators of partial differential equations on a diffeomorphic family of domains.
\newblock \emph{arXiv preprint arXiv:2402.07250}, 2024.

\bibitem[Yin et~al.(2022)Yin, Kirchmeyer, Franceschi, Rakotomamonjy, and Gallinari]{yin2022continuous}
Yuan Yin, Matthieu Kirchmeyer, Jean-Yves Franceschi, Alain Rakotomamonjy, and Patrick Gallinari.
\newblock Continuous pde dynamics forecasting with implicit neural representations.
\newblock \emph{arXiv preprint arXiv:2209.14855}, 2022.

\bibitem[Yu et~al.(2022)Yu, Tang, Rao, Huang, Zhou, and Lu]{yu2022point}
Xumin Yu, Lulu Tang, Yongming Rao, Tiejun Huang, Jie Zhou, and Jiwen Lu.
\newblock Point-bert: Pre-training 3d point cloud transformers with masked point modeling.
\newblock In \emph{Proceedings of the IEEE/CVF conference on computer vision and pattern recognition}, pages 19313--19322, 2022.

\bibitem[Zhao et~al.(2025)Zhao, Liu, Li, Chen, and Liu]{zhao2025DNO}
Zhiwei Zhao, Changqing Liu, Yingguang Li, Zhibin Chen, and Xu~Liu.
\newblock Diffeomorphism neural operator for various domains and parameters of partial differential equations.
\newblock \emph{Communications Physics}, 8\penalty0 (1):\penalty0 15, 2025.

\end{thebibliography}

\appendix
\section{Convergence Study}
The corresponding latent mesh resolutions for the results in Figure~\ref{fig:convergence} are presented in Table~\ref{tab:resoluion-settings}.
\begin{table}[h]
    \centering
    \begin{tabular}{ccccc}
        \toprule
        Model/ Sampling rate & 100\% & 62\% & 27\% & 16\%\\
        \midrule
        GeoFNO & [40,40,40] & [35,35,35] & [26,26,26] & [22,22,22]\\ 
        GINO & [64,64,64] & [54,54,54] & [42,42,42] & [36,36,36] \\
        OTNO(ours) & [104,104] & [80,80] & [54,54] & [42,42]\\
        \bottomrule
    \end{tabular}
    \caption{Latent Resolution Settings for Convergence Analysis: This table details the resolution configurations in latent space for each model at different sampling rates. Optimal expansion factors were applied for each method to enhance convergence. 
    }
    \label{tab:resoluion-settings}
\end{table}

For GeoFNO(3D), the latent mesh is set to be the sphere surface; for GeoFNO(2D), the latent mesh is a square. As shown in Table~\ref{tab:geofno-dimension-comparison}, GeoFNO(3D) outperforms GeoFNO(2D) obviously, therefore we only present the GeoFNO(3D) in Figure~\ref{fig:convergence}.
\begin{table}[h]
    \centering
    \begin{tabular}{lcccc}
        \toprule
        Model/ Sampling Rate & 16\% & 27\% & 62\% & 100\%\\
        \midrule
        GeoFNO(2D) & 0.2593 & 0.2351 & 0.2147 & 0.1626 \\
        GeoFNO(3D) & 0.2193 & 0.2102 & 0.1458 & 0.1350\\
        \bottomrule
    \end{tabular}
    \caption{Comparison of Different Latent Space Dimension for GeoFNO}
    \label{tab:geofno-dimension-comparison}
\end{table}

\section{DrivAerNet: Drag Coefficient}
\label{appendix: Cd}
On DrivAerNet dataset, our target prediction is drag coefficient which is a good metric to evaluate a car design. For a fluid with unit density, the drag coefficient is defined as
\begin{equation}
c_d = \frac{2}{v^2 A} \left( \int_{\partial\Omega} p(\mathbf{x}) (\mathbf{\hat{n}}(\mathbf{x}) \cdot \mathbf{\hat{i}}(\mathbf{x})) \, d\mathbf{x} + \int_{\partial\Omega} \mathbf{T}_w(\mathbf{x}) \cdot \mathbf{\hat{i}}(\mathbf{x}) \, d\mathbf{x} \right)    
\end{equation}

where $\partial\Omega \subset \mathbb{R}^3$ is the surface of the car, $p: \mathbb{R}^3 \to \mathbb{R}$ is the pressure, $\mathbf{\hat{n}}: \partial\Omega \to S^2$ is the outward unit normal vector of the car surface, $\mathbf{\hat{i}}: \mathbb{R}^3 \to S^2$ is the unit direction of the inlet flow, $\mathbf{T}_w: \partial\Omega \to \mathbb{R}^3$ is the wall shear stress on the surface of the car, $v \in \mathbb{R}$ is the speed of the inlet flow, and $A \in \mathbb{R}$ is the area of the smallest rectangle enclosing the front of the car.

In many practical situations involving high Reynolds number flows over streamlined bodies (like airfoils), the pressure drag is significantly higher than the shear drag. Given that DrivAerNet Reynold number is roughly  $9.39 \times 10^6$
which is high, we only consider the first term, the pressure drag term, in the above drag coefficient formula. Thus we design the loss function in our model as follows:

\begin{equation}
    Loss = \sum_{j=1}^{N} ( \sum_{i}^n p(\mathbf{x}_i) (\mathbf{\hat{n}}(\mathbf{x}_i) \cdot \mathbf{\hat{i}}(\mathbf{x}_i)) - (c_d)_j )^2.
\end{equation}


\section{FlowBench}
\label{appendix: flowbench}
\begin{figure}[h]
    \centering
    \begin{minipage}{1.0\textwidth}
        \centering
        \includegraphics[width=\linewidth]{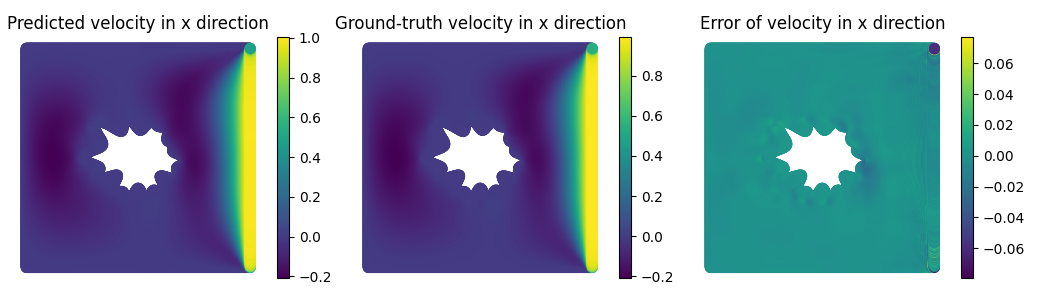}
    \end{minipage}\\
    \begin{minipage}{1.0\textwidth}
        \centering
        \includegraphics[width=\linewidth]{JMLR/figures/fullspace_vy.png}
    \end{minipage}\\
    \begin{minipage}{1.0\textwidth}
        \centering
        \includegraphics[width=\linewidth]{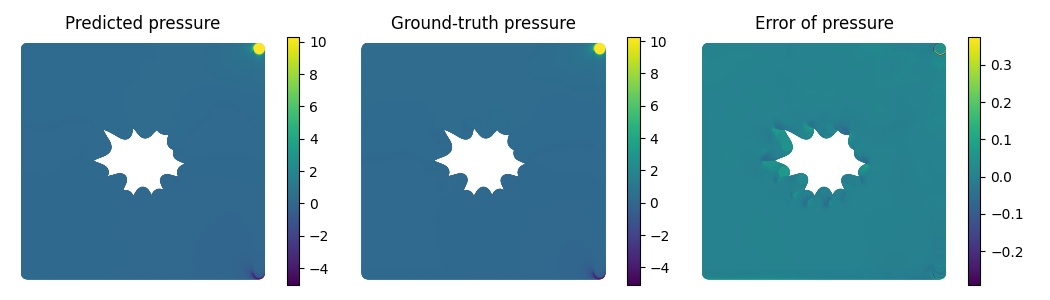}
    \end{minipage}\\
    \caption{Results Visualization of Flow Dataset under M1 metric (full space)}
    \label{fig:flowbench_fullspace}
\end{figure}

\begin{figure}[h]
    \centering
    \begin{minipage}{1.0\textwidth}
        \centering
        \includegraphics[width=\linewidth]{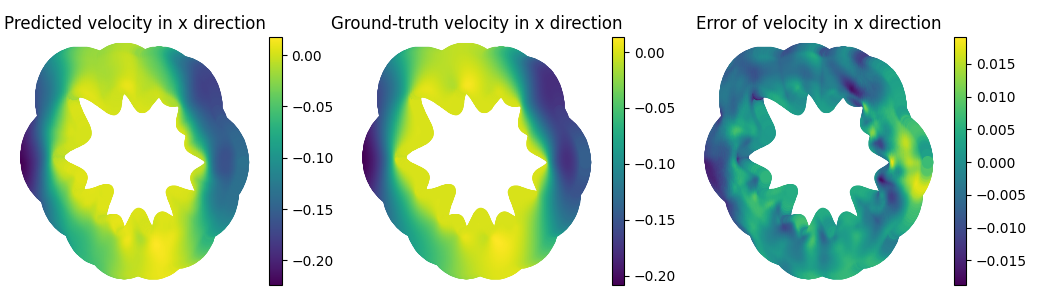}
    \end{minipage}\\
    \begin{minipage}{1.0\textwidth}
        \centering
        \includegraphics[width=\linewidth]{JMLR/figures/boundary_vy.png}
    \end{minipage}\\
    \begin{minipage}{1.0\textwidth}
        \centering
        \includegraphics[width=\linewidth]{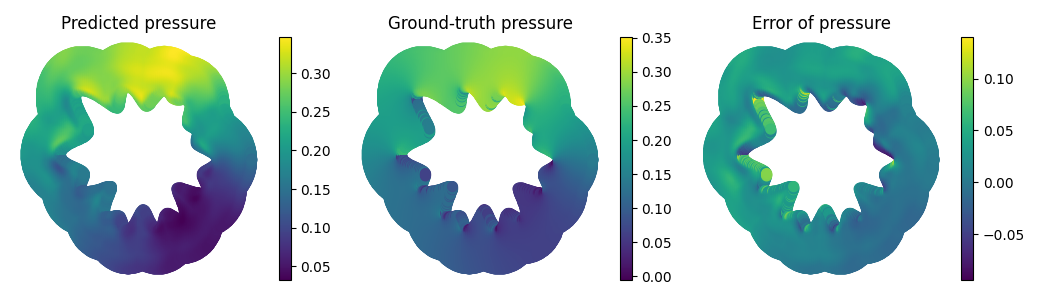}
    \end{minipage}\\
    \caption{Results Visualization of Flow Dataset under M2 metric (boundary)}
    \label{fig:flowbench_boundary}
\end{figure}

The inputs include Reynolds number, geometry mask (a binary representation of shape), and signed distance function (SDF); outputs are velocity in the x and y directions and pressure. Since our model is designed for geometry operator learning, we opt to test it on the Easy case, which uses a standard 80-20 random split, unlike the Hard case, which splits the dataset according to Reynolds number. We evaluate our model across three groups within the dataset: G1, which consists of parametric shapes generated using Non-Uniform Rational B-Splines (NURBS) curves; G2, which consists of parametric shapes generated using spherical harmonics; and G3, which consists of non-parametric shapes sampled from the grayscale dataset in SkelNetOn. For each group, the diversity of shapes is significantly greater than that of car designs in the aforementioned car dataset. The complete visualization of results are as shown in Fig.~\ref{fig:flowbench_fullspace} and Fig.~\ref{fig:flowbench_boundary}

\end{document}